\documentclass{article}
\usepackage{microtype}
\usepackage{graphicx}
\usepackage{booktabs}
\usepackage{array}
\usepackage{hyperref}
\usepackage{url}
\usepackage{xurl}
\usepackage{amsmath}
\usepackage{amssymb}
\usepackage{amsthm}
\usepackage{capt-of}
\usepackage{subcaption}
\usepackage{xcolor}
\usepackage[most]{tcolorbox}
\usepackage{varwidth}

\usepackage[accepted]{mlsys2025}

\usepackage{amsmath,amsfonts,bm}

\def\eqref#1{equation~\ref{#1}}

\def\1{\bm{1}}

\DeclareMathAlphabet{\mathsfit}{\encodingdefault}{\sfdefault}{m}{sl}
\SetMathAlphabet{\mathsfit}{bold}{\encodingdefault}{\sfdefault}{bx}{n}

\allowdisplaybreaks

\graphicspath{{imgs/}}

\newcommand{\method}{CLiS}
\newcommand{\methodfull}{\underline{C}ertified \underline{L}oop-\underline{i}nvariant \underline{S}tabilization}
\newcommand{\DiTReductionMax}{19.6\%}
\newcommand{\AttentionReductionMax}{24.0\%}
\definecolor{queryorange}{HTML}{E69F00}
\definecolor{keyblue}{HTML}{0072B2}
\definecolor{takeawayframe}{HTML}{293021}
\definecolor{takeawayback}{HTML}{F2F6FC}
\definecolor{headerboxbg}{HTML}{EFF4F9}
\newcommand{\qidx}{\textcolor{queryorange}{i}}
\newcommand{\kidx}{\textcolor{keyblue}{j}}
\newcommand{\qvar}[1]{\textcolor{queryorange}{#1}}
\newcommand{\kvvar}[1]{\textcolor{keyblue}{#1}}
\newtcolorbox{takeawaybox}[1]{
  enhanced,
  colback=takeawayback,
  colframe=takeawayframe,
  colbacktitle=takeawayframe,
  coltitle=white,
  boxrule=0.55pt,
  arc=1mm,
  left=1.2mm,
  right=1.2mm,
  top=1.8mm,
  bottom=0.8mm,
  before skip=8pt,
  after skip=4pt,
  fontupper=\normalsize,
  fonttitle=\bfseries\normalsize,
  title={#1},
  attach boxed title to top left={xshift=1.2mm,yshift=-1mm},
  varwidth boxed title*=-4mm,
  boxed title style={
    colframe=takeawayframe,
    colback=takeawayframe,
    boxrule=0pt,
    arc=0.8mm,
    left=1mm,
    right=1mm,
    top=0.45mm,
    bottom=0.45mm
  }
}

\newtcolorbox{headerbox}{
  enhanced,
  colback=headerboxbg,
  frame hidden,
  boxrule=0pt,
  arc=3.5mm,
  left=6mm,
  right=6mm,
  top=5.5mm,
  bottom=5.5mm,
  before skip=0pt,
  after skip=10pt
}

\mlsystitlerunning{CLiS: Plug-and-Play Attention Acceleration for Video Diffusion on TPUs via Certified Stabilization}

\hypersetup{
  pdftitle={CLiS: Plug-and-Play Attention Acceleration for Video Diffusion on TPUs via Certified Stabilization},
  pdfauthor={Elisa Tsai, Shao-yu Huang, Daniel Mandragona, Wei-Yu Lin, WenXin Dong, Jevin Jiang, Chandra Devarakonda, Mehdy Bohlool, Yixin Shi}
}

\begin{document}

\twocolumn[
\thispagestyle{empty}
\begin{headerbox}
{\raggedright\sffamily\bfseries\fontsize{17.5}{21.5}\selectfont
CLiS: Plug-and-Play Attention Acceleration for Video Diffusion on TPUs via Certified Stabilization\par}
\vspace{9pt}
{\raggedright\sffamily\bfseries\normalsize
\mbox{Elisa Tsai\textsuperscript{1*$\dagger$}}\hspace{1.1em}%
\mbox{Shao-yu Huang\textsuperscript{1*}}\hspace{1.1em}%
\mbox{Daniel Mandragona\textsuperscript{1}}\hspace{1.1em}%
\mbox{Wei-Yu Lin\textsuperscript{1}}\hspace{1.1em}%
\mbox{WenXin Dong\textsuperscript{1}}\hspace{1.1em}%
\mbox{Jevin Jiang\textsuperscript{1}}\hspace{1.1em}%
\mbox{Chandra Devarakonda\textsuperscript{1}}\hspace{1.1em}%
\mbox{Mehdy Bohlool\textsuperscript{1}}\hspace{1.1em}%
\mbox{Yixin Shi\textsuperscript{1}}\par}
\vspace{5pt}
{\raggedright\rmfamily\small
\textsuperscript{1}Google\hspace{1.4em}\textsuperscript{*}Equal contribution\hspace{1.4em}\textsuperscript{$\dagger$}Corresponding author\par}
\vspace{9pt}
{\rmfamily\normalsize\noindent
High-resolution video diffusion relies on attention over long token sequences,
making attention efficiency critical to generation latency. Our analysis
identifies repeated maximum reductions and accumulator rescaling in online
softmax as a bottleneck in long-context TPU attention. We propose \method{}
(\methodfull{}), a training-free, plug-and-play optimization for video diffusion
inference. \method{} fixes per-row stabilizers before the KV loop and certifies
numerical safety per head, falling back to online softmax when certification
fails. It integrates into existing fused-attention kernels without changing
model weights or architecture. On Wan~2.2 and LTX-2 across TPU~v6e and TPU7x,
\method{} reduces diffusion transformer (DiT) step latency by up to \DiTReductionMax{} relative to tuned
online softmax without measurable quality loss. Controlled TPU adaptations of
prior GPU stabilization methods show performance or numerical-safety
limitations. Preliminary causal LLM experiments also show up to 13.5\% higher
training throughput in continued pretraining and 7.7\% in supervised fine-tuning.\par}
\vspace{8pt}
{\raggedright\rmfamily\small
\textcolor{mydarkblue}{\{\href{mailto:elisatsai@google.com}{elisatsai}, \href{mailto:shyhuang@google.com}{shyhuang}, \href{mailto:dandragona@google.com}{dandragona}, \href{mailto:weiyulin@google.com}{weiyulin}, \href{mailto:wenxindong@google.com}{wenxindong}, \href{mailto:jevinjiang@google.com}{jevinjiang}, \href{mailto:chandrasekhard@google.com}{chandrasekhard}, \href{mailto:mehdy@google.com}{mehdy}, \href{mailto:yxshi@google.com}{yxshi}\}@google.com}\par}
\end{headerbox}

\begin{center}
  \includegraphics[width=0.9\textwidth]{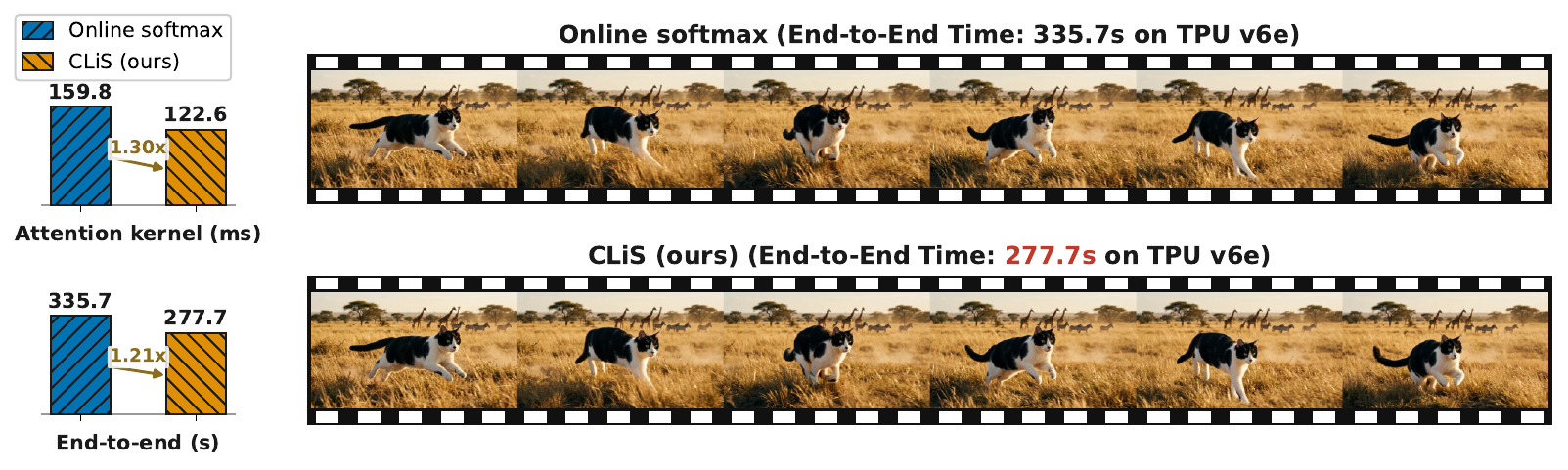}
  \captionof{figure}{\method{} accelerates long-sequence video diffusion attention without
  visually altering generation quality. Left: speedup on TPU~v6e. Right: frames 0, 20, 40, 60, 80, 100 from a 720p, 7.5s (16fps, 40-step) video generated by WAN 2.2~\citep{wan2025}.}
  \label{fig:teaser}
\end{center}

\vskip 0.1in
]

\section{Introduction}
\label{sec:intro}

\begin{figure}
  \centering
  \includegraphics[width=0.8\linewidth]{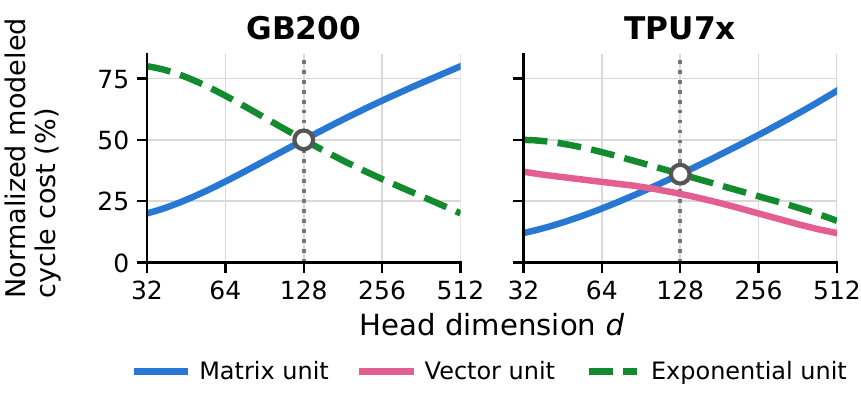}\\[2pt]
  \includegraphics[width=\linewidth]{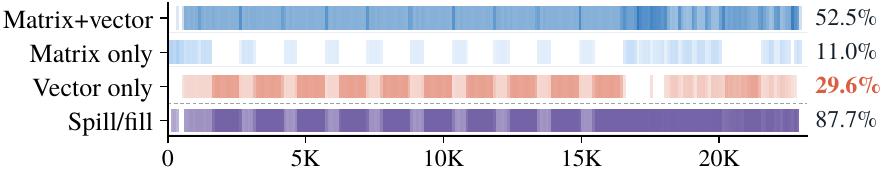}
  \vspace{-6pt}
  \caption{\textbf{(1)} Each unit's modeled share of cycles per attention score
  (work divided by peak throughput). At $d{=}128$ (dotted line, used by Wan~2.2
  and LTX-2), exponentiation alone matches the matmuls. On TPU7x, vector-unit max
  reduction and rescaling push softmax past them.
  \textbf{(2)} Exposed softmax in the compiled VLIW bundle schedule for Wan~2.2
  DiT self-attention (one KV-block iteration) on TPU7x.}
  \label{fig:matmul_exp_throughput_imbalance}
\end{figure}

Video Diffusion Transformers (DiTs) repeatedly apply self-attention over
spatiotemporal tokens during generation~\citep{peebles2023scalable,wan2025,kong2024hunyuanvideo,hacohen2024ltx,brooks2024sora,deepmind2025veo}.
As resolution and duration increase, these sequences become extremely
long: a five-second 720p video can require approximately 75K--120K
tokens~\citep{wan2025,kong2024hunyuanvideo}. Attention efficiency is
therefore a major determinant of generation latency.
FLOP counts, however, can obscure an emerging hardware bottleneck.
Although matmuls dominate attention FLOPs, softmax can rival them in cycle cost because exponential throughput is far lower (\autoref{fig:matmul_exp_throughput_imbalance}) and has scaled more slowly~\citep{shah2024flashattention3}.

\textbf{Motivation.}
This matmul--non-matmul throughput imbalance, previously studied on GPUs~\citep{shah2024flashattention3,zadouri2026flashattention4}, also arises on TPUs. 
Schedules from the TPU compiler's low-level optimizer (LLO), which emits the final VLIW instruction bundles~\citep{jouppi2021ten}, show that exponentiation alone rivals the matmuls in modeled cycle cost, while maximum reduction and accumulator rescaling push the full softmax cost beyond them. 
TPU architecture adds another constraint: within each KV iteration, online-softmax state
competes with score and probability tiles for a limited set of shared vector registers.
Because this maximum-and-rescaling chain is re-executed for every KV block, its total cost
grows with sequence length, and runtime ablations confirm that it remains on the critical
path (Section~\ref{sec:softmax_analysis}).

\textbf{Challenges.}
Softmax subtracts a per-row \emph{stabilizer} from the logits to keep exponentials in
floating-point range. In exact arithmetic, any choice gives the same result.
Online softmax uses the largest logit seen so far, so every KV block that raises it
forces a rescale of the accumulated sums.
A \emph{loop-invariant} stabilizer, fixed before the KV loop, removes both the per-block
maximum reduction and the rescaling.
Prior GPU work moves partway toward this: FlashDecoding++ fixes a shared shift but
recomputes rows that violate it~\citep{hong2024flashdecodingpp}, VFA freezes an approximate
maximum without a safety guarantee~\citep{sun2026vfa}, and FlashAttention-4 rescales only
past a threshold but still computes every block's maximum~\citep{zadouri2026flashattention4}.
Yet if a fixed stabilizer sits too far below a row's maximum, exponentials overflow, and if
it sits too far above, they underflow to zero (\autoref{fig:exp_safety_ranges}).
Computing the exact maximum in advance avoids both but costs an extra pass over all
attention scores.
The challenge is therefore to \textit{certify a safe loop-invariant stabilizer at
negligible cost}.

\textbf{Our Method.}
We introduce \method{} (\methodfull{}), a training-free, plug-and-play
optimization for existing video diffusion models. It replaces online softmax's
running maximum with a loop-invariant per-row stabilizer whenever numerical
safety can be certified. 
Using inexpensive query and key statistics, \method{} derives an upper bound on each row's attention logits before the key-block loop. 
Certified heads skip maximum reduction and accumulator rescaling on every KV block, while uncertified heads fall back to standard online softmax. 
By changing only attention stabilization, \method{} integrates into existing
fused-attention kernels without additional training or changes to model weights
or architecture. It preserves standard attention in exact arithmetic.

\textbf{Results.}
We evaluate \method{} on Wan~2.2 and LTX-2 using TPU~v6e and TPU7x, varying video resolution and duration over attention lengths from 17K to 184K tokens. 
Relative to tuned online softmax, \method{} reduces attention-kernel latency by up to \AttentionReductionMax{} and DiT-step latency by up to \DiTReductionMax{}, without measurable degradation in
generation quality. 
TPU adaptations of prior GPU methods either fail to deliver speedups or produce invalid outputs. 
A preliminary exploration of causal LLM training also finds up to $13.5\%$ higher
training throughput in continued pretraining and $7.7\%$ in supervised fine-tuning
(Section~\ref{sec:pretraining}), suggesting potential applicability beyond diffusion attention.

\textbf{Contributions.}

\noindent\textbf{(1)} We identify repeated maximum reduction and accumulator
rescaling as a bottleneck in long-context diffusion attention on latest-generation TPUs.

\noindent\textbf{(2)} We propose \method{}, a training-free, plug-and-play method
that removes repeated stabilization through per-head numerical certification.

\noindent\textbf{(3)} We achieve up to \textbf{\DiTReductionMax{}} lower DiT-step
latency on Wan~2.2 and LTX-2 without measurable quality loss. Preliminary LLM
experiments also show speedups in continued pretraining and supervised fine-tuning.

\section{Preliminaries}
\label{sec:background}

\subsection{FlashAttention and Online Softmax}

Given queries, keys, and values
$Q,K,V\in\mathbb{R}^{N\times d}$, scaled dot-product attention computes
\begin{equation}
    S = \frac{QK^\top}{\sqrt{d}}, \qquad
    P = \operatorname{softmax}(S), \qquad
    O = PV,
\end{equation}
where $N$ is the sequence length and $d$ is the head dimension.
FlashAttention partitions $Q$ into query blocks $\{Q_i\}$ and $K,V$ into
key-value blocks $\{K_j,V_j\}$, avoiding materialization of the
$N\times N$ matrices $S$ and $P$~\citep{dao2022flashattention}.

For each query row, online softmax maintains a running maximum $m$, a
normalization accumulator $\ell$, and an output accumulator $o$.
Given the logits $s^{(j)}$ for key block $j$, it updates
\begin{align}
    \widehat{m}^{(j)}
        &= \max_k s_k^{(j)}, \\
    m^{(j)}
        &= \max\!\left(m^{(j-1)},\widehat{m}^{(j)}\right), \\
    \alpha^{(j)}
        &= \exp\!\left(m^{(j-1)}-m^{(j)}\right), \\
    \ell^{(j)}
        &= \alpha^{(j)}\ell^{(j-1)}
         + \sum_k \exp\!\left(s_k^{(j)}-m^{(j)}\right), \\
    o^{(j)}
        &= \alpha^{(j)}o^{(j-1)}
         + \sum_k \exp\!\left(s_k^{(j)}-m^{(j)}\right)v_k^{(j)}.
\end{align}
The final attention output is $o^{(T)}/\ell^{(T)}$.

\begin{figure}
  \centering
  \includegraphics[width=0.8\linewidth]{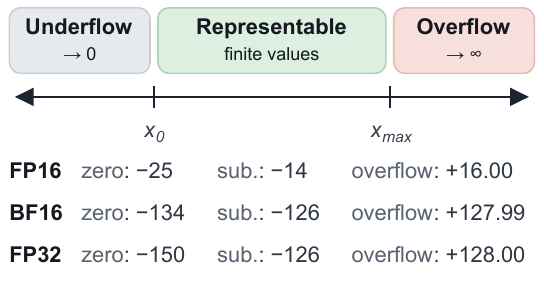}
  \caption{Approximate numerical limits of $\operatorname{exp2}(x)=2^x$ for FP16, BF16, and FP32~\citep{kalamkar2019bfloat16}.}
  \label{fig:exp_safety_ranges}
\end{figure}

\textbf{Numerically Stable Softmax}. Softmax is invariant to a uniform shift:
\begin{equation}
    \operatorname{softmax}(s)
    =
    \operatorname{softmax}\!\left(s-c\,\mathbf{1}_N\right)
\end{equation}
for any scalar $c$ in exact arithmetic.
Conventional online softmax chooses the running row maximum as this shift,
ensuring that every exponent is non-positive~\citep{milakov2018online}.
As shown in~\autoref{fig:exp_safety_ranges}, in floating-point arithmetic, replacing the running maximum with an arbitrary fixed shift can cause overflow, underflow, or increased numerical
error~\citep{blanchard2021accurately}.

\subsection{TPU Architecture and Low-Level Execution}
\label{sec:tpu_architecture}

Attention combines matrix multiplications with reductions, exponentials, and elementwise updates.
These operations run on different TPU units, so performance depends on how their execution overlaps.

\begin{figure}[t!]
\centering
\includegraphics[width=0.8\linewidth]{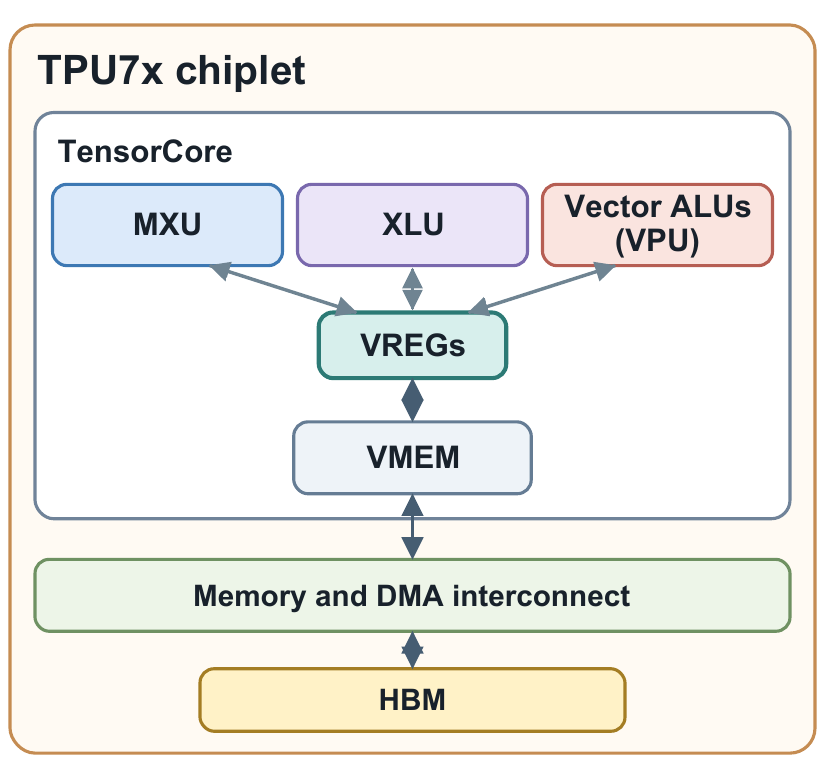}
\caption{Simplified TPU7x chiplet architecture and dataflow for
attention~\citep{google2025tpuv7x,jax2026tpudetails}. Arrows show logical dataflow.
Each TPU7x chip contains two chiplets.}
\label{fig:tpu_architecture}
\end{figure}

\textbf{Compute units.}
A TPU's main compute core, called a \emph{TensorCore}, contains specialized units
for these operations. As shown in~\autoref{fig:tpu_architecture},
the Matrix Multiply Unit (MXU) computes the two attention matrix products,
$QK^\top$ and $PV$. The Vector Processing Unit (VPU) performs elementwise
operations, including exponentiation and accumulator rescaling. The Cross-Lane
Unit (XLU) supports data movement across vector lanes, which combines with vector
operations to implement row-wise reductions. These units can execute concurrently
when their data dependencies permit~\citep{jouppi2021ten,jax2026tpudetails}.

\textbf{Registers and memory.}
These computations share vector registers (VREGs), which hold intermediate
results such as score tiles, probabilities, and online-softmax accumulators.
On-chip vector scratchpad memory (VMEM) provides additional storage, while
off-chip high-bandwidth memory (HBM) holds the larger input and output tensors.
When live intermediate values exceed VREG capacity, the compiler temporarily
stores them in VMEM (\emph{VREG spills}) and reloads them into VREGs when needed
(\emph{fills})~\citep{jax2026tpudetails}. We refer to these transfers as
\emph{spill/fill traffic}.

\begin{figure*}[!t]
  \centering
  \includegraphics[width=\textwidth]{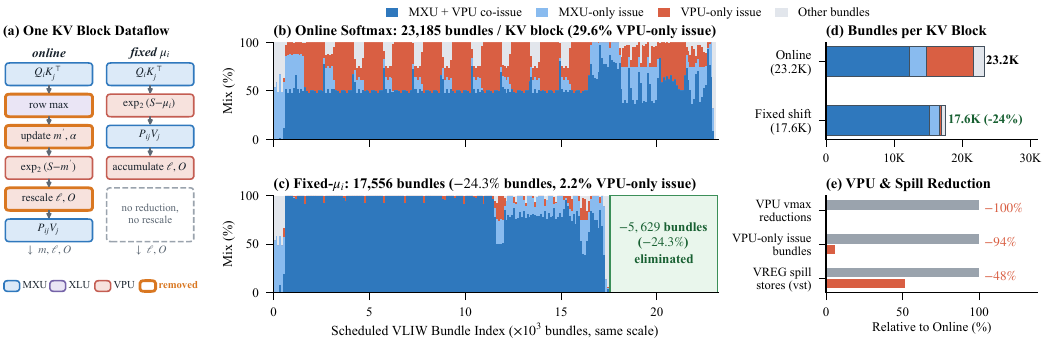}
  \vspace{-6pt}
  \caption{\textbf{Online-stabilization overhead for one compiled KV-block iteration.}
  Wan~2.2 on TPU7x, generating a 5\,s, 720p video (81~frames), at identical tile sizes.
  \textbf{(a)} A loop-invariant stabilizer $\mu_i$ removes the reduction and rescaling chain.
  \textbf{(b--c)} Compiled VLIW schedules. Fixed stabilization reduces VPU-only issue bundles and shortens the schedule by $24.3\%$.
  \textbf{(d--e)} Scheduled bundles, VPU \texttt{vmax} reductions, VPU-only issue bundles, and VREG spill stores. Counts describe instruction issue, not elapsed unit-active cycles.}
  \label{fig:tpu_attention_critical_path}
\end{figure*}

\textbf{Instruction scheduling.}
Pallas, a JAX kernel language, lowers TPU kernels through Mosaic to LLO, a
hardware-specific representation for TPU compilation~\citep{jax2026pallasdesign}.
The compiler coordinates the functional units using a \emph{very long instruction
word} (VLIW) execution model. It groups instructions into \emph{bundles}:
instructions for different units that can be issued together~\citep{jouppi2021ten}.
A bundle can therefore issue matrix work alongside independent vector work.
We perform LLO analysis of the compiled instruction bundles and their
generation-specific modeled cycle costs to examine where matrix and vector work
overlap, where spill/fill instructions occur, and where dependencies prevent further MXU
work from being issued.

\section{Motivation: What Online Softmax Costs on TPUs}
\label{sec:softmax_analysis}

Online softmax repeats a row-maximum reduction and accumulator rescaling for
every KV block. We first examine this stabilization cost, then distinguish the
effects of sequence length and tile selection, and finally identify the numerical
constraint on removing it.

\subsection{Online Stabilization Adds Removable Vector Work}
\label{sec:motivation_cost}

We compare compiled schedules and run latency ablations to isolate the cost of
online stabilization. We replace the running maximum with a loop-invariant
stabilizer, removing the maximum reduction and accumulator rescaling while
retaining exponentiation. \autoref{fig:tpu_attention_critical_path} shows the
removed dependency chain.

We examine WAN 2.2 ($d=128$)~\citep{wan2025} generating a 5-second 720p video.
At identical tile sizes, we compare one compiled KV-block iteration.
The compiled schedule shrinks from $23{,}185$ to $17{,}556$ instruction
bundles ($24.3\%$). The fixed stabilizer also cuts spill stores by $48\%$.
Bundles issuing VPU work without an MXU instruction fall from $6{,}858$ to
$384$ ($94\%$).

In runtime ablations on this workload, removing maximum reduction and rescaling reduces DiT-step
latency by $7.5\%$. Removing all softmax work reduces it by $8.2\%$. 
Therefore, stabilization accounts for $91\%$ of the removable softmax overhead
in this ablation.

\begin{takeawaybox}{Takeaway 1: Stabilization is a major removable cost.}
Maximum reduction and accumulator rescaling account for most removable softmax
overhead in the measured workload. A fixed per-row stabilizer removes both while
retaining exponentiation.
\end{takeawaybox}

\subsection{Sequence Length and Tile Size Have Different Effects}
\label{sec:motivation_length}

\begin{figure*}[t!]
\centering

\begin{subfigure}[t]{0.49\textwidth}
\centering
\includegraphics[width=\textwidth]{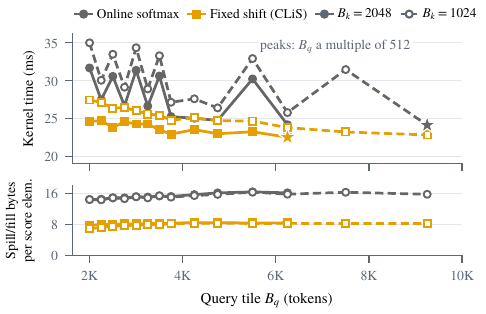}
\caption{Query-tile sweep at $N=75{,}600$ (720p, 81~frames).
Stars mark fastest tiles. Spill/fill traffic is reported in bytes per attention score.}
\label{fig:context_tiling_qtile}
\end{subfigure}
\hfill
\begin{subfigure}[t]{0.49\textwidth}
\centering
\includegraphics[width=\textwidth]{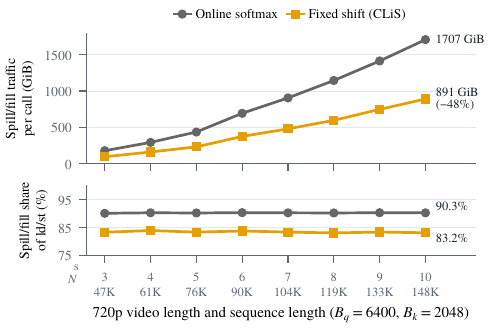}
\caption{Sequence-length sweep at fixed tiles (720p).
Top: total spill/fill traffic per call. Bottom: spill/fill share of
vector loads and stores.}
\label{fig:context_tiling_seqlen}
\end{subfigure}

\caption{\textbf{Loop-invariant stabilization reduces spill/fill traffic across tile sizes and sequence lengths.}
One ring attention step of WAN~2.2 on TPU7x
(20 heads, $d=128$, BF16). Times are medians of 20 warm calls. Spill/fill traffic is derived
from static LLO counts.}
\label{fig:context_tiling_spills}

\end{figure*}

To separate the effects of tile size and sequence length on stabilization
overhead, we compare online softmax with a loop-invariant stabilizer in WAN~2.2
on TPU7x (\autoref{fig:context_tiling_spills}).

\textbf{Tile selection.}
In Pallas, the TPU compiler maps configurable query and key tiles to vector
registers and VMEM and schedules work across execution
units~\citep{jax2026pallasdesign,jax2026tpudetails}. We therefore tune tile
sizes empirically to balance data reuse, memory demand, and scheduling
efficiency.

At a fixed sequence length (\autoref{fig:context_tiling_qtile}), larger query
tiles increase spill/fill traffic per KV block, but traffic per score stays
nearly constant. A loop-invariant stabilizer roughly halves this per-score
traffic across all tiles. Online softmax also exhibits latency peaks at certain
tile sizes that persist after removing spill/fill code, indicating a scheduling
cost beyond memory traffic. A loop-invariant stabilizer avoids these peaks,
providing more consistent latency and a $6.6\%$ speedup at each method's
fastest tile.

\textbf{Sequence length.}
At fixed tile sizes (\autoref{fig:context_tiling_seqlen}), longer sequences add
more query--key tile pairs. Because each pair incurs similar spill/fill
traffic, total traffic grows with sequence length while remaining a nearly
constant share of vector loads and stores. A loop-invariant stabilizer roughly
halves this traffic across all measured lengths.

\begin{takeawaybox}{Takeaway 2: Loop-invariant stabilization reduces spill/fill traffic and tile sensitivity.}
A loop-invariant stabilizer roughly halves spill/fill traffic per score and
reduces latency sensitivity to tile choice, whereas longer sequences increase
total traffic by adding query--key tile pairs.
\end{takeawaybox}

\subsection{Removing Stabilization Requires Numerical Safety}
\label{sec:motivation_challenge}

Removing row-maximum reduction and accumulator rescaling speeds up attention,
but can introduce numerical errors. Although softmax is shift-invariant in
exact arithmetic, the choice of stabilizer matters in floating point.
As shown in~\autoref{fig:exp_safety_ranges}, a stabilizer too far below the row maximum can cause exponentiation to
overflow. One too far above it can cause underflow.

Simple alternatives each have limitations. Omitting the shift risks overflow,
while clipping changes the attention probabilities. A shared stabilizer must
accommodate different logit ranges across rows and heads. Computing each row's
exact maximum avoids this difficulty but requires an extra pass over the
attention scores. We therefore need a \textit{per-row stabilizer that is safe,
inexpensive to compute, and fixed before the KV loop}.

\section{Certified Loop-Invariant Stabilization}
\label{sec:method}

Certification is performed independently for each attention head.
All statistics below refer to one head, and we suppress the head index unless
needed. Processing multiple heads in parallel does not combine their certificates.

Following FlashAttention's block notation~\citep{dao2022flashattention}, let
$\qvar{Q_i}$ be a query block and $(\kvvar{K_j},\kvvar{V_j})$ a key--value
block. Orange denotes query-indexed variables and blue denotes key--value-indexed
variables, while pairwise arrays retain colored indices. We absorb the attention
scale and $\log_2 e$ into $Q$, so the score block is
$S_{\qidx\kidx}=\qvar{Q_i}\kvvar{K_j}^{\top}$ and softmax uses \texttt{exp2}.

\paragraph{Bounding and centering.}
Let $N_k$ denote the number of valid key tokens, excluding padding.
We compute their mean $\bar{k}=\frac{1}{N_k}\sum_{\kidx=1}^{N_k}\kvvar{k_j}$
and subtract it from each key, giving
$\kvvar{\widetilde{k}_j}=\kvvar{k_j}-\bar{k}$.
For each query $\qvar{q_i}$, this subtracts the same scalar
$\qvar{q_i}^{\top}\bar{k}$ from every logit in its row and preserves
attention~\citep{zhang2025sageattention}. In exact arithmetic, the centered
logits in each row have mean zero, so their
maximum $\qvar{a_i}=\max_{\kidx}\qvar{q_i}^{\top}
\kvvar{\widetilde{k}_j}$ satisfies $\qvar{a_i}\geq 0$.  Let
\begin{equation}
  \kappa=\max_{\kidx}\lVert\kvvar{\widetilde{k}_j}\rVert_2,
  \qquad
  \qvar{B_i}=\left\lceil\lVert\qvar{q_i}\rVert_2\,\kappa\right\rceil .
  \label{eq:clis-bound}
\end{equation}
By the Cauchy--Schwarz inequality, $\qvar{a_i}\leq\qvar{B_i}$. For an integer recentering
constant $C$, \method{} uses the loop-invariant shift
\begin{equation}
  \qvar{\mu_i}=\qvar{B_i}-C.
  \label{eq:clis-shift}
\end{equation}
Using $\qvar{B_i}$ directly as the stabilizer can cause underflow when the
bound is loose. We therefore subtract a constant $C$, raising every exponent
argument by $C$. This allows positive exponent arguments while keeping them
bounded above by $C$. We choose $C$ subject to the numerical safety conditions
below to avoid an all-zero denominator and prevent overflow.

\paragraph{Numerical certificate.}
Before using the loop-invariant path, we check that each row retains a
nonzero denominator and that exponentiation and accumulation stay within
range. Let $[L_{\exp},U_{\exp}]$ be a conservative input interval over which
\texttt{exp2} produces nonzero, finite outputs. Let $U_{\mathrm{acc}}$ be the
safe FP32 accumulator magnitude limit expressed in base-2 logarithmic units.
We require $\qvar{B_i}$ to bound the computed logits, including rounding
error. The supplied margin $\rho$ must guarantee that each computed centered
row has maximum at least $-\rho$. Our implementation uses $\rho=1$.
For head $h$, let $\qvar{B_{h,r}}$ denote the bound for query row $r$.
We define $B_h=\max_{1\leq r\leq N_q}\qvar{B_{h,r}}$, taking the maximum
only over rows within that head. The value-magnitude bound
$G=\lceil\log_2\max(1,\max_{j,r}|V_{jr}|)\rceil$ also uses that head's values.
We seek a nonnegative integer recentering constant $C$ satisfying
\begin{equation}
  \begin{gathered}
  C-B_h-\rho\geq L_{\exp},\quad C\leq U_{\exp},\\
  C+\lceil\log_2 N_k\rceil+G<U_{\mathrm{acc}}.
  \end{gathered}
  \label{eq:clis-certificate}
\end{equation}
If a solution exists, we choose the largest feasible integer $C$ and use the
loop-invariant path. Otherwise, the head uses online softmax.
The first condition keeps at least one exponential in each row nonzero,
so the denominator remains positive. The second prevents individual
exponentials from overflowing. The third reserves enough range to sum over
$N_k$ keys and accumulate the value-weighted outputs. For eligible heads,
the fixed per-row shift
preserves attention in exact arithmetic. \hyperref[alg:clis]{Algorithm~\ref*{alg:clis}} summarizes
this dispatch.

\begin{algorithm}[!t]
\caption{\method{} forward pass for one dense attention head.}\label{alg:clis}
\small
\noindent\textbf{Inputs (one head):} Finite $Q\in\mathbb R^{N_q\times d}$,
$K\in\mathbb R^{N_k\times d}$, $V\in\mathbb R^{N_k\times d_v}$.
Positive integer $N_q,N_k,B_q,B_k$, with all keys valid.
$Q$ includes the attention scale and $\log_2 e$.
Supply $L_{\exp},U_{\exp},U_{\mathrm{acc}}$. Set $\rho\gets1$.
\textsc{BoundUp} evaluates Equation~\ref{eq:clis-bound} conservatively.

\medskip
\noindent\begin{tabular}{@{}r@{\hspace{0.45em}}>{\raggedright\arraybackslash}p{0.88\linewidth}@{}}
1 & $\bar{k}\gets N_k^{-1}\sum_{r=1}^{N_k}\kvvar{k_r}$. \\
2 & $\widetilde K\gets K-\mathbf1\bar{k}^{\top}$. \\
3 & $\kappa\gets$ a conservative upper bound on $\max_r\|\kvvar{\widetilde k_r}\|_2$. \\
4 & $B_h\gets\max_{1\leq r\leq N_q}\textsc{BoundUp}(\qvar{q_r},\kappa)$. \\
5 & $G\gets\lceil\log_2\max(1,\max_{j,r}|\kvvar{V_{jr}}|)\rceil$. \\
6 & $A\gets\lceil\log_2N_k\rceil+G$. \\
7 & $C\gets\min\bigl(\lfloor U_{\exp}\rfloor,\lceil U_{\mathrm{acc}}-A\rceil-1\bigr)$. \\
8 & \textbf{if} $C<\max(0,B_h+\rho+L_{\exp})$ \textbf{then return} \textsc{OnlineExp2}$(Q,K,V)$. \\
9 & Partition $Q$ into $T_r=\lceil N_q/B_q\rceil$ blocks $\qvar{Q_i}$ with row sets $\mathcal I_i$. \\
10 & Partition $(\widetilde K,V)$ into $T_c=\lfloor N_k/B_k\rfloor$ full blocks and a tail of width $B_\tau=N_k-T_cB_k$. \\
11 & Allocate $O\in\mathbb R^{N_q\times d_v}$. \\
12 & \textbf{for} $\qidx=1,\ldots,T_r$ \textbf{do} \\
13 & \quad $\qvar{B_i}\gets[\textsc{BoundUp}(\qvar{q_r},\kappa)]_{r\in\mathcal I_i}$. \\
14 & \quad $\qvar{\mu_i}\gets \qvar{B_i}-C\mathbf1$. \\
15 & \quad $\qvar{O_i}\gets0\in\mathbb R^{|\mathcal I_i|\times d_v}$, $\qvar{\ell_i}\gets0\in\mathbb R^{|\mathcal I_i|}$. \\
16 & \quad \textbf{for} $\kidx=1,\ldots,T_c$ \textbf{do} \\
17 & \qquad $S_{\qidx\kidx}\gets \qvar{Q_i}\kvvar{\widetilde K_j}^{\top}$. \\
18 & \qquad $P_{\qidx\kidx}\gets\operatorname{exp2}(S_{\qidx\kidx}-\qvar{\mu_i}\mathbf1_{B_k}^{\top})$. \\
19 & \qquad $\qvar{\ell_i}\gets\qvar{\ell_i}+\operatorname{rowsum}(P_{\qidx\kidx})$. \\
20 & \qquad $\qvar{O_i}\gets \qvar{O_i}+P_{\qidx\kidx}\kvvar{V_j}$. \\
21 & \quad \textbf{end for} \\
22 & \quad \textbf{if} $B_\tau>0$ \textbf{then} \\
23 & \qquad $S_{\qidx\tau}\gets \qvar{Q_i}\kvvar{\widetilde K_\tau}^{\top}$. \\
24 & \qquad $\qvar{m_i}\gets\max\bigl(\qvar{\mu_i},\operatorname{rowmax}(S_{\qidx\tau})\bigr)$. \\
25 & \qquad $\qvar{\alpha_i}\gets\operatorname{exp2}(\qvar{\mu_i}-\qvar{m_i})$. \\
26 & \qquad $P_{\qidx\tau}\gets\operatorname{exp2}(S_{\qidx\tau}-\qvar{m_i}\mathbf1_{B_\tau}^{\top})$. \\
27 & \qquad $\qvar{\ell_i}\gets\qvar{\alpha_i}\odot\qvar{\ell_i}+\operatorname{rowsum}(P_{\qidx\tau})$. \\
28 & \qquad $\qvar{O_i}\gets\operatorname{diag}(\qvar{\alpha_i})\qvar{O_i}+P_{\qidx\tau}\kvvar{V_\tau}$. \\
29 & \quad \textbf{end if} \\
30 & \quad Write $O[\mathcal I_i,:]\gets\operatorname{diag}(\qvar{\ell_i})^{-1}\qvar{O_i}$ to HBM. \\
31 & \textbf{end for} \\
32 & \textbf{return} $O$. \\
\end{tabular}

\end{algorithm}

\paragraph{Compiler-aware dispatch.}
We precompute each head's key mean $\bar{k}$, norm bound $\kappa$, and
eligibility, reducing across sequence shards when needed. The kernel computes
$\qvar{B_i}$ and $\qvar{\mu_i}$ once per query block while $\qvar{Q_i}$ is
on chip. Each head selects its execution path before the KV loop, keeping
eligibility checks outside it.

For Mosaic scheduling, both paths use an online update on any final partial
KV block, rescaling existing accumulators as needed to preserve the result
in exact arithmetic. Certified heads perform row-maximum reduction only for
this block.

\section{Experiments}
\label{sec:experiments}

\begin{figure*}[t!]
  \centering
  \includegraphics[width=\linewidth]{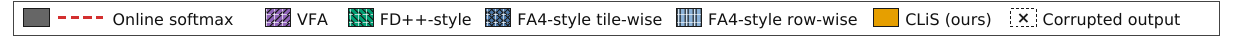}\par\vspace{3pt}\nointerlineskip

  \begin{subfigure}{\textwidth}
    \includegraphics[width=\linewidth]{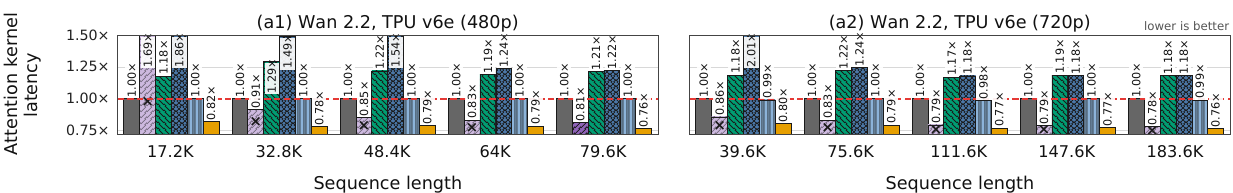}
    \phantomsubcaption\label{fig:main_eval_wan22_v6e_kernel}
  \end{subfigure}\par\vspace{-8pt}\nointerlineskip
  \begin{subfigure}{\textwidth}
    \includegraphics[width=\linewidth]{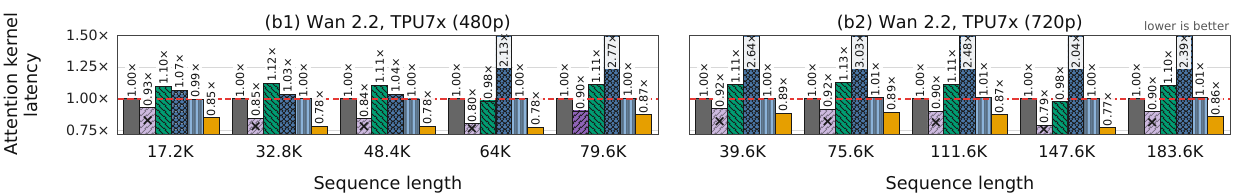}
    \phantomsubcaption\label{fig:main_eval_wan22_v7x_kernel}
  \end{subfigure}\par\vspace{-8pt}\nointerlineskip
  \begin{subfigure}{\textwidth}
    \includegraphics[width=\linewidth]{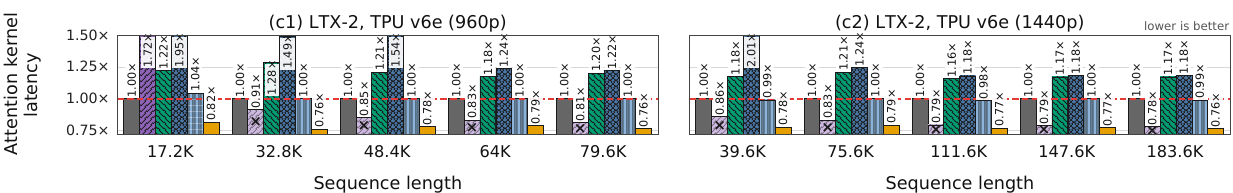}
    \phantomsubcaption\label{fig:main_eval_ltx2_v6e_kernel}
  \end{subfigure}\par\vspace{-8pt}\nointerlineskip
  \begin{subfigure}{\textwidth}
    \includegraphics[width=\linewidth]{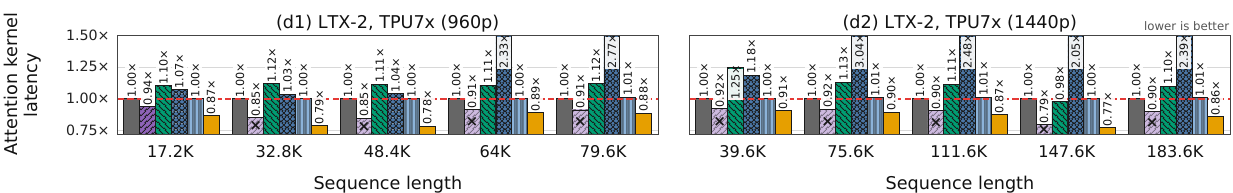}
    \phantomsubcaption\label{fig:main_eval_ltx2_v7x_kernel}
  \end{subfigure}

  \caption{Normalized attention-kernel latency for \method{} and baselines,
  relative to online softmax (1.0$\times$). Lower is better.
  All runs use eight TensorCores with CFG$=$2$\times$Ring$=$4 (Ulysses$=$1).
  Panels (a1)--(d2) use focused axes. Out-of-range bars are clipped and
  labeled with their measured ratios. Crosses mark corrupted outputs.}
  \label{fig:main_eval_all}
\end{figure*}

\begin{figure*}[t!]\ContinuedFloat
  \centering
  \includegraphics[width=\linewidth]{imgs/main_eval/main_eval_legend.pdf}\par\vspace{3pt}\nointerlineskip

  \begin{subfigure}{\textwidth}
    \includegraphics[width=\linewidth]{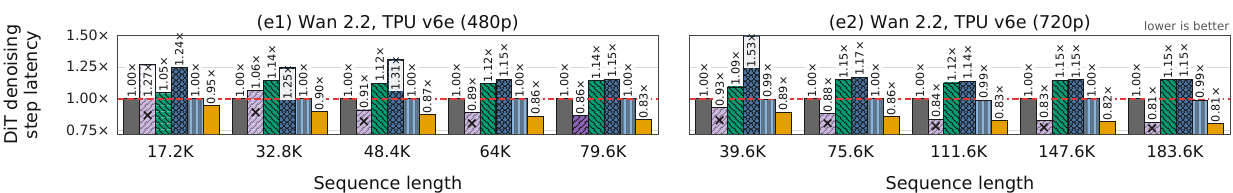}
    \phantomsubcaption\label{fig:main_eval_wan22_v6e}
  \end{subfigure}\par\vspace{-8pt}\nointerlineskip
  \begin{subfigure}{\textwidth}
    \includegraphics[width=\linewidth]{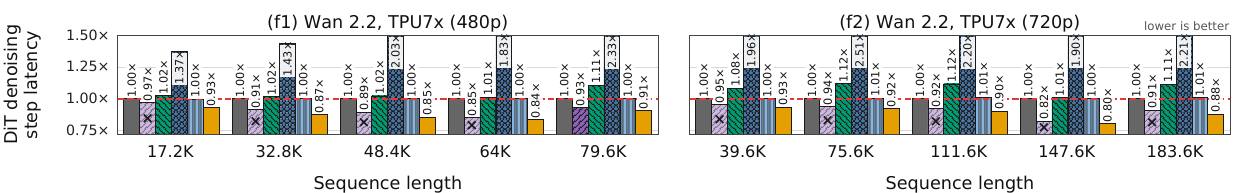}
    \phantomsubcaption\label{fig:main_eval_wan22_v7x}
  \end{subfigure}\par\vspace{-8pt}\nointerlineskip
  \begin{subfigure}{\textwidth}
    \includegraphics[width=\linewidth]{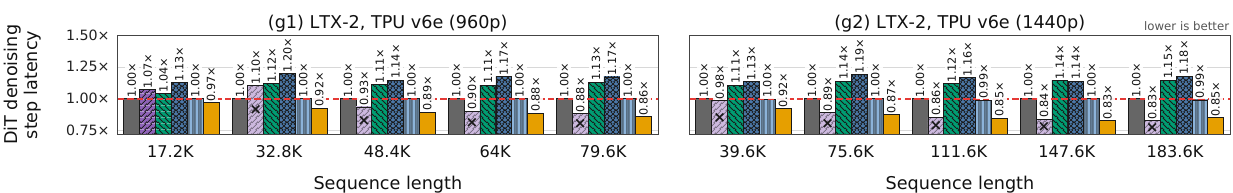}
    \phantomsubcaption\label{fig:main_eval_ltx2_v6e}
  \end{subfigure}\par\vspace{-8pt}\nointerlineskip
  \begin{subfigure}{\textwidth}
    \includegraphics[width=\linewidth]{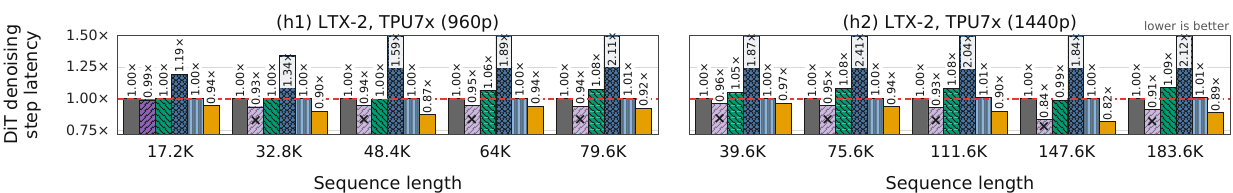}
    \phantomsubcaption\label{fig:main_eval_ltx2_v7x}
  \end{subfigure}

  \caption{Normalized DiT-step latency (continued).
  Panels (e1)--(h2) follow the same order and encoding as panels (a1)--(d2).}
\end{figure*}

In this section, we evaluate \method{} on two video diffusion models, Wan~2.2~\citep{wan2025} and LTX-2~\citep{hacohen2026ltx2}, across TPU~v6e~\citep{google2024tpuv6e} and TPU7x~\citep{jouppi2026tpuv7x}.
Our results show that \method{} consistently accelerates attention and diffusion inference over tuned online softmax while preserving numerical accuracy and generation quality, whereas competing methods either provide no speedup or produce invalid outputs.

\subsection{Experimental Setup}

\textbf{Baselines.}
We compare \method{} with the following baselines:
1) \textbf{Online softmax}~\citep{milakov2018online,dao2022flashattention}: A tuned, open-sourced Pallas~\citep{pallas2023} implementation of the standard running-maximum algorithm.
2) \textbf{FlashDecoding++-style (FD++)}~\citep{hong2024flashdecodingpp}: Uses an empirically calibrated shared shift and recomputes an affected row with online softmax when the supported exponent range is violated.
3) \textbf{VFA-style}~\citep{sun2026vfa}: Initializes each row's maximum from approximate key-block representatives, updates it on sink and local blocks, and freezes it for the remaining blocks.
4) \textbf{FlashAttention-4-style (FA4)}~\citep{zadouri2026flashattention4}: Keeps the exact per-block row maximum and rescales only when its increase exceeds the threshold.  FA4 decides per row but coalesces correction over 32-row warps. Our port compares row-wise and full-tile control and reports both.

The latter three methods include GPU-specific scheduling and dataflow
optimizations that do not transfer directly to TPUs. Our \emph{style} baselines
adapt their stabilization mechanisms, including safety checks and recovery
behavior, within a common Pallas kernel.
See Appendix~\ref{app:baselines}.

\textbf{Implementations.}
Wan~2.2 and LTX-2 use TorchAX~\citep{torchax}.
All workloads use eight TensorCores (TPU~v6e-8 or TPU7x-8), sharding
classifier-free guidance (CFG) and ring attention as CFG$=$2$\times$Ring$=$4
(Ulysses$=$1). Appendix~\ref{app:parallelism_ablation} varies this layout.
For each workload, we tune each variant's tile sizes under otherwise matched
execution settings. Reported latency includes preprocessing, safety checks,
and recomputation. We measure attention-kernel and DiT-step latency after
warmup using repeated runs.

\textbf{Experiments.}
We vary video duration at each fixed resolution to evaluate how \method{}
speedup changes with sequence length. For Wan~2.2, we use 41, 81, 121, 161,
and 201 frames (2.5, 5, 7.5, 10, and 12.5\,s). For LTX-2, we use 81, 161,
241, 321, and 401 frames (3.4, 6.7, 10.0, 13.4, and 16.7\,s
at 24\,fps).

\subsection{Kernel and DiT-Step Performance}

\autoref{fig:main_eval_all} compares attention-kernel and DiT-step latency
across both models, TPU generations, and sequence lengths. All results are
normalized to the corresponding online-softmax control, with lower values
indicating lower latency. As shown in Tables~\ref{tab:kernel_latency_absolute}
and~\ref{tab:dit_latency_absolute} in Appendix~\ref{app:absolute_latency},
we also report the corresponding absolute latencies.

\textbf{Attention-kernel latency.}
\method{} reduces kernel latency by $9.4$--\AttentionReductionMax{} across all measured
workloads. The benefit appears on both Wan~2.2 and LTX-2 and on both TPU~v6e
and TPU7x, showing that it extends beyond a single model or accelerator.
The reduction varies across configurations and does not grow monotonically
with sequence length.

\textbf{DiT-step latency.}
The kernel improvements translate into $3.1$--\DiTReductionMax{} lower DiT-step
latency. The smaller reductions reflect the other computation in each step,
which attention optimization does not accelerate. Longer sequences often
show larger step-level gains, but the trend varies by model and TPU generation.

\textbf{Comparison with prior stabilization mechanisms.}
The FlashDecoding++-style and FlashAttention-4-style baselines provide little
benefit or increase latency in most configurations. Row-wise FA4 remains
close to online softmax, while full-tile correction incurs larger overhead.
VFA can reduce latency but produces corrupted outputs on most Wan~2.2 and LTX-2 runs (9 of 10 workloads each).
Latency and output quality must therefore be considered together when comparing
stabilization methods.

\subsection{Certification and Output Quality}

\begin{table*}[t!]
\begin{center}

\caption{\textbf{Safety and output quality on Wan~2.2.}
Certification and attention error cover ten TPU7x workloads (17K--184K tokens).
\emph{Valid} counts workloads without NaN/Inf or all-zero outputs.
Safe-condition rates use different units and are not directly comparable.
The \method{} rate spans the five 720p workloads, from the same run as
Table~\ref{tab:fallback_accounting} and Figure~\ref{fig:fallback_scaling}.
Errors and VBench deltas are relative to online softmax.
VBench uses 110 prompts with matched seeds at 720p$\times$81f.
VFA videos have all-zero frames and are unscored.}
\vspace{0.2cm}
\label{tab:safety_quality}
\small
\setlength{\tabcolsep}{4.0pt}
\resizebox{\textwidth}{!}{%
\begin{tabular}{l c c c c c c c c c}
\toprule
  & \multicolumn{2}{c}{Certification}
  & \multicolumn{2}{c}{Attention output}
  & \multicolumn{5}{c}{$\Delta$VBench vs.\ online ($\rightarrow 0$)} \\
\cmidrule(lr){2-3}
\cmidrule(lr){4-5}
\cmidrule(lr){6-10}
Method
  & Valid
  & Safe-cond.\ rate
  & Max abs.
  & Rel.\ $\ell_2$
  & Imaging
  & Subject cons.
  & Temp.\ flicker
  & Motion smooth.
  & Dynamic degree \\ \midrule
Online softmax & 10/10 & -- & 0 & 0 & 0 & 0 & 0 & 0 & 0 \\
\midrule
VFA-style & 1/10 & 40.1 & $3.1\!\times\!10^{-2}$ & $2.53\!\times\!10^{-3}$ & -- & -- & -- & -- & -- \\
FD++-style & 10/10 & 91.0 & $3.1\!\times\!10^{-2}$ & $2.73\!\times\!10^{-3}$ & $+0.0011$ & $-0.0005$ & $+0.0012$ & $-0.0001$ & $-0.1000$ \\
FA4-style (tile) & 10/10 & -- & $3.1\!\times\!10^{-2}$ & $2.05\!\times\!10^{-3}$ & $+0.0032$ & $+0.0013$ & $-0.0001$ & $+0.0010$ & 0 \\ \midrule
\method{} & \textbf{10/10} & 97.9--98.3 & $3.1\!\times\!10^{-2}$ & $2.7\!\times\!10^{-3}$ & $+0.0110$ & $+0.0015$ & $+0.0004$ & $+0.0002$ & 0 \\
\bottomrule
\end{tabular}
}

\end{center}

\end{table*}

\begin{table*}[t!]
\begin{center}

\caption{Certified-head eligibility and the cost of the fallback path.
Wan~2.2 T2V, $1280\!\times\!720$, 81 frames, 75{,}600 tokens.}
\vspace{0.2cm}
\label{tab:fallback_accounting}
\footnotesize

\begin{subtable}[t]{0.63\textwidth}
\centering
\caption{Eligibility and maximum-reduction work removed.}
\setlength{\tabcolsep}{3pt}
\begin{tabular}{l c c c c c}
\toprule
  {Expert} & Eligible $\uparrow$ & Fallback $\downarrow$ & Sites
  & Max-reduce $\downarrow$ & Score rows $\downarrow$ \\ \midrule
High-noise & 98.54\% & 1.46\% & 36 / 1600 & 6.65\% & 3.90\% \\
Low-noise  & 98.07\% & 1.93\% & 48 / 1600 & 7.09\% & 4.36\% \\ \midrule
Total      & \textbf{98.22\%} & \textbf{1.78\%} & 84 / 3200 & \textbf{6.95\%} & \textbf{4.21\%} \\
\bottomrule
\end{tabular}
\end{subtable}
\hfill
\begin{subtable}[t]{0.35\textwidth}
\centering
\caption{Had the 84 fallback sites run fixed-$m$.}
\setlength{\tabcolsep}{3pt}
\begin{tabular}{l c c}
\toprule
  {Outcome} & Max logit & Sites \\ \midrule
No effect          & $\leq 108$   & 60 \\
Graceful flush     & $108$--$122$ & 1 \\
NaN                & $> 122$      & \textbf{23} \\
\bottomrule
\end{tabular}
\end{subtable}

\end{center}

\end{table*}

\begin{figure*}[t!]
\centering

\begin{subfigure}[t]{0.49\textwidth}
\centering
\includegraphics[width=\textwidth]{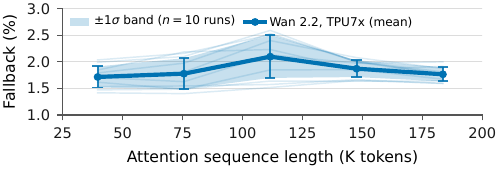}
\caption{Fallback rate across attention sequence lengths.}
\label{fig:fallback_vs_length}
\end{subfigure}
\hfill
\begin{subfigure}[t]{0.49\textwidth}
\centering
\includegraphics[width=\textwidth]{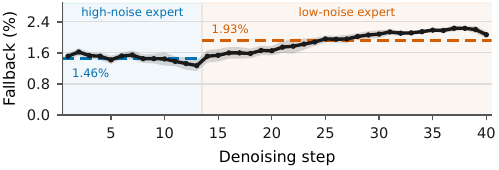}
\caption{Fallback rate across denoising steps.}
\label{fig:fallback_vs_step}
\end{subfigure}

\caption{Runtime fallback rate of \method{} on Wan~2.2 (TPU7x) as a function of sequence length (left) and denoising step (right).}
\label{fig:fallback_scaling}

\end{figure*}

\textbf{Numerical safety.}
Table~\ref{tab:safety_quality} shows that \method{} produces valid outputs
on all ten tested Wan~2.2 workloads on TPU7x, as do the FlashDecoding++-style
and FA4-style baselines. VFA-style fails on nine workloads, highlighting
the risk of using an uncertified stabilizer without a recovery mechanism.
\method{} preserves attention in exact arithmetic. Its measured attention
output differs from online softmax by $2.7\!\times\!10^{-3}$ in relative
$\ell_2$, comparable to valid baselines.

\textbf{Generation quality.}
We evaluate the generated videos using selected VBench~\citep{huang2024vbench} metrics with matched
prompts and seeds. \method{} shows small differences from online softmax
across the reported dimensions, supporting that its attention-level numerical
differences do not materially degrade generation quality in this evaluation.
Full VBench scores, paired differences, and visual comparisons appear in
Appendix~\ref{app:safety_measurements}
(Tables~\ref{tab:vbench_absolute} and~\ref{tab:vbench_delta_ci}, and
Figure~\ref{fig:vbench_visual_comparison}).

\subsection{Fallback Behavior}
\label{sec:fallback}

\paragraph{Fallback is rare.}
As shown in~\autoref{tab:fallback_accounting}a, $98.22\%$ of head evaluations
use loop-invariant stabilization and $1.78\%$ fall back to online softmax
in the 40-step Wan~2.2 run at 720p with 81 frames.
As shown in~\autoref{fig:fallback_scaling}, fallback remains near $2\%$
across sequence lengths and denoising steps.

\paragraph{Fallback prevents numerical failures.}
As shown in~\autoref{tab:fallback_accounting}b, replaying the 84 rejected
head sites without fallback would produce NaNs at 23 sites and flush a ring
shard's weights to zero at one site. The remaining 60 sites would be
numerically harmless in this replay. The certificate is therefore
conservative, but it catches cases that require fallback for numerical safety.

\paragraph{Most maximum reductions are eliminated.}
As shown in~\autoref{tab:fallback_accounting}a, the hybrid implementation
removes $93.05\%$ of \texttt{reduce\_max} operations and $95.79\%$ of score-row
reductions relative to online softmax. Therefore, the small fallback fraction
still allows most of the repeated maximum-reduction work to be removed.

\subsection{Preliminary Exploration of Causal LLM Training}
\label{sec:pretraining}

\begin{table*}[t!]
\centering
\caption{Representative causal LLM pretraining, continual pretraining, and SFT results on TPU7x (64 chips, or 512 chips for DSv3-671B). Forward attention kernel latency change, training speedup (step throughput or wall-clock time-to-train, TTT), and final-50 loss are paired against online softmax.}
\vspace{0.1cm}
\label{tab:pretraining_summary}
\footnotesize
\setlength{\tabcolsep}{4.5pt}
\begin{tabular*}{\textwidth}{@{\extracolsep{\fill}}l l c c c c c c@{}}
\toprule
Model (Dataset) & Task & Seq. & Fwd. Attn. & Train Speedup & Loss (Base / \method{}) & $\Delta$ Loss & Cert. \\
\midrule
Qwen3-8B (C4)            & Pretrain                       & 8K   & $-37.9\%$ & $+7.8\%$       & 3.182 / 3.185 & $+0.10\%$ & 100\% \\
Qwen3-8B (ClimbMix)      & Pretrain                       & 8K   & $-37.9\%$ & $+8.9\%$       & 2.678 / 2.673 & $-0.18\%$ & 100\% \\
Qwen3-8B (UltraChat)     & SFT                            & 8K   & $-36.2\%$ & $+7.7\%$       & 0.807 / 0.806 & $-0.03\%$ & 95.8\% \\
Qwen3-8B (PG-19)         & Continual PT                   & 256K & $-35.6\%$ & $+12.4\%$      & 2.182 / 2.182 & $-0.02\%$ & 95.5\% \\
\midrule
Llama-3.1-8B (UltraChat) & SFT                            & 8K   & $-37.8\%$ & $+6.7\%$       & 0.752 / 0.752 & $-0.07\%$ & 100\% \\
Llama-3.1-8B (PG-19)     & Continual PT                   & 128K & $-43.3\%$ & $+11.6\%$      & 1.992 / 1.992 & $-0.03\%$ & 99.8\% \\
Llama-3.1-8B (PG-19)     & Continual PT                   & 256K & $-43.6\%$ & $+13.5\%$      & 2.013 / 2.013 & $-0.01\%$ & 99.8\% \\
\midrule
Gemma-3-12B (C4)         & Pretrain                       & 8K   & $-9.6\%$  & $+0.6\%$       & 3.423 / 3.425 & $+0.06\%$ & 100\% \\
DSv3-671B (MLPerf)       & Continual PT (loss $\le 3.60$) & 4K   & $-22.1\%$ & $+6.2\%$ (TTT) & 3.589 / 3.593 & $+0.11\%$ & 99.9\% \\
DSv3-12L (Ring MLA)      & Pretrain                       & 128K & $-33.4\%$ & $+19.0\%$      & 10.70 / 10.70 & $\phantom{+}0.00\%$ & 100\% \\
\bottomrule
\end{tabular*}
\end{table*}

\begin{figure*}[t!]
\centering
\begin{minipage}[t]{\columnwidth}
\centering
\includegraphics{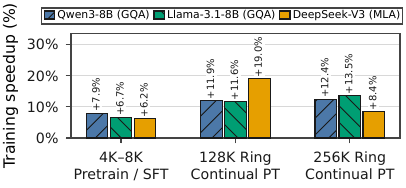}
\caption{End-to-end training speedup of \method{} over online softmax across sequence lengths from 4K to 256K tokens on TPU7x.}
\label{fig:pretraining_speedup}
\end{minipage}\hfill
\begin{minipage}[t]{\columnwidth}
\centering
\includegraphics[width=\columnwidth]{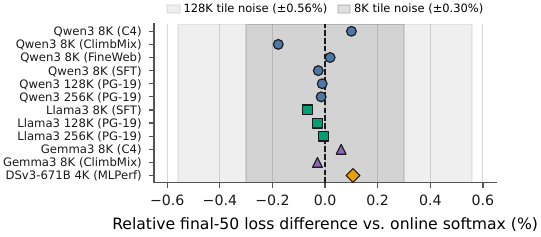}
\caption{Relative final-50 training loss difference of \method{} against online softmax, compared with hardware tile-size reduction noise.}
\label{fig:pretraining_loss_delta}
\end{minipage}
\end{figure*}

To explore applicability beyond diffusion attention, we conduct preliminary
experiments with causal Splash and Ring Attention for autoregressive LLM
training on TPU7x. The forward kernel uses loop-invariant
stabilization for certified heads and stores a compact log-sum-exp table for
the unchanged backward pass.

We evaluate four model families across 8K scratch pretraining for 1{,}500 steps
on C4~\citep{raffel2020t5}, ClimbMix~\citep{diao2025climb}, and
FineWeb-Edu~\citep{penedo2024fineweb};
8K supervised fine-tuning on UltraChat~\citep{ding2023ultrachat};
128K--256K continual pretraining with Ring Attention on
PG-19~\citep{rae2020compressive}; and 512-chip FP8 MLA training of
DeepSeek-V3-671B~\citep{deepseekai2024deepseekv3}.

\paragraph{Training throughput.}
\autoref{fig:pretraining_speedup} and \autoref{tab:pretraining_summary} show
that \method{} accelerates end-to-end training steps by $6.7\%$ to $8.9\%$ at
8K context on Qwen3-8B~\citep{yang2025qwen3} and
Llama-3.1-8B~\citep{grattafiori2024llama3}. At 128K and 256K context, where Ring
Attention accounts for a larger share of step time, \method{} reduces forward
attention kernel latency by $33.4\%$ to $43.6\%$ and improves end-to-end step
throughput by $11.6\%$ to $19.0\%$. On 512-chip DeepSeek-V3-671B, \method{}
reduces forward Splash kernel time by $22.1\%$ and cuts wall-clock
time-to-train to the MLPerf validation target~\citep{mlcommons2026training}
by $6.2\%$. In Gemma-3-12B~\citep{gemmateam2025gemma3}, five
of every six layers use 1{,}024-token sliding windows, so applying \method{} to
the global 8K layers yields a modest $+0.6\%$ step speedup.

\paragraph{Loss convergence and certificate behavior.}
As shown in \autoref{fig:pretraining_loss_delta}, final-50 training loss under
\method{} remains within $-0.18\%$ to $+0.11\%$ of online softmax across all
certified runs. This variation is smaller than the $\pm 0.30\%$ to $\pm 0.56\%$
loss shift caused by changing the online-softmax KV reduction block size from
512 to 256. Models with bounded query-key norms, either from architectural
QK-Norm (Qwen3-8B and Gemma-3-12B) or from pretrained checkpoints (Llama-3.1-8B
and DeepSeek-V3), maintain $95.5\%$ to $100\%$ head certification throughout
training. In contrast, training Llama-3.1-8B from random initialization without
QK-Norm allows unconstrained logit growth past the FP32 exponent budget by step
40, causing the certificate to fall back to online softmax and prevent NaNs.

\section{Related Work}
\label{sec:related_work}

\paragraph{Exact attention kernels.}
FlashAttention and FlashAttention-2 optimize IO and work
partitioning~\citep{dao2022flashattention,dao2024flashattention2}.
FlashAttention-3 and FlashAttention-4 optimize matrix and softmax schedules for
Hopper and Blackwell~\citep{shah2024flashattention3,zadouri2026flashattention4}.
These kernels retain blockwise maximum reductions. Ragged Paged Attention uses
Pallas for variable-length TPU LLM serving~\citep{jiang2026ragged}.

\paragraph{Reducing softmax overhead.}
Online softmax retains a running maximum~\citep{milakov2018online}.
FlashDecoding++ uses calibrated shared shifts with recovery~\citep{hong2024flashdecodingpp}, while
VFA predicts and freezes maxima~\citep{sun2026vfa}.
FlashAttention-4 skips small rescalings, while FLASH-D uses a normalization
recurrence~\citep{zadouri2026flashattention4,alexandridis2025flash}.
TurboAttention quantizes and approximates exponentials~\citep{kang2025turboattention}.
\method{} certifies fixed per-row shifts, removing loop-level maximum reductions
and rescaling with online-softmax fallback.

\paragraph{Video diffusion attention.}
SageAttention uses low-bit quantization, while SpargeAttention adds
sparsity~\citep{zhang2025sageattention,zhang2024sageattention2,
zhang2026sageattention3,zhang2025spargeattention}. BLASST skips negligible
blocks~\citep{yuan2026blasst}. Sliding Tile Attention,
Sparse VideoGen, and Radial Attention exploit spatiotemporal
sparsity~\citep{zhang2025fast,xi2025sparsevideogen,li2025radialattention}.
Sparse VideoGen2 and HyperVAttention use
clustering~\citep{yang2025sparsevideogen2,lee2026hypervattention}.
\method{} complements these methods while preserving dense attention in exact arithmetic.

\section{Conclusion}
\label{sec:conclusion}

We presented \method{}, a training-free, plug-and-play method that removes
repeated online-softmax stabilization work in TPU attention. It uses certified
loop-invariant stabilizers when safe and otherwise falls back to online softmax.
\method{} integrates into existing attention kernels without changing model
weights or architecture. On Wan~2.2 and LTX-2 across TPU~v6e and TPU7x,
\method{} reduces attention-kernel latency by up to \AttentionReductionMax{}
and DiT-step latency by up to \DiTReductionMax{} relative to tuned online softmax,
without measurable generation-quality loss. Preliminary causal LLM experiments
also show up to 13.5\% higher end-to-end training throughput in continued
pretraining and 7.7\% in supervised fine-tuning.

\bibliography{references}
\bibliographystyle{mlsys2025}

\clearpage
\appendix

\twocolumn[{
\begin{minipage}{\textwidth}
  \centering
  \footnotesize
  \setlength{\tabcolsep}{4pt}
  \captionof{table}{Controlled comparison of the stabilization mechanisms and
  their TPU adaptations. ``Block'' denotes one streamed KV block.
  \method{} still computes a row maximum and rescales on the final partial KV
  block, if one exists (Algorithm~\ref{alg:clis}, lines 22--28).
  $^\ast$FD++ fast path only, with full-row online recomputation after a range
  violation. $^\dagger$Certified heads only. Rejected heads use online softmax.}
  \label{tab:appendix_baseline_comparison}
  \begin{tabular}{p{0.13\textwidth}p{0.19\textwidth}p{0.09\textwidth}p{0.13\textwidth}p{0.29\textwidth}}
    \toprule
    Method & Retained stabilizer & Row max & Rescale rule & Controlled TPU adaptation \\
    \midrule
    Online softmax
      & Exact running maximum
      & Every & Every
      & Tuned exact reference in the common kernel \\
    FD++-style
      & Calibrated shared shift
      & No$^\ast$ & No$^\ast$
      & Range check and exact row recomputation without flat-GEMM or CUDA scheduling \\
    VFA-style
      & Predicted maximum, then frozen
      & Sink/local only & Sink/local only
      & Locality mapped to flattened video tokens, including representative construction \\
    FA4-style
      & Running maximum with thresholded updates
      & Every & $\Delta_j>8$
      & Row-wise and full-tile control without Blackwell pipelining or exp emulation \\
    \method{}
      & Certified fixed shift
      & No$^\dagger$ & No$^\dagger$
      & Per-head certification with unchanged online fallback \\
    \bottomrule
  \end{tabular}

\end{minipage}
\vspace{\dbltextfloatsep}
}]

\section{Baseline Selection and TPU Adaptations}
\label{app:baselines}

Our baselines are controlled ports of softmax-side mechanisms rather than
end-to-end reproductions of GPU systems.  Every method uses the same Pallas attention kernel, tiling, sharding,
masking, precision, and ring schedule~\citep{liu2024ringattention}. Only the stabilization rule and its
required checks or preprocessing change.  This design attributes differences
to the mechanism under study rather than to a different matmul pipeline, tile,
or communication schedule.

\subsection{Why These Baselines}

Tuned online softmax is the primary reference because it is exact, robust, and
differs from \method{} only in its stabilization logic.  FlashDecoding++, VFA,
and FlashAttention-4 provide three distinct ways to reduce that logic: use a
calibrated shift with recovery, predict a maximum and later freeze it, or keep
the exact running maximum but skip small rescalings, respectively
\citep{hong2024flashdecodingpp,sun2026vfa,zadouri2026flashattention4}.  These
baselines test whether a heuristic plus recovery, approximate maximum
prediction, or conditional rescaling is sufficient without certification.
For each \emph{style} baseline, we preserve the published softmax decision and
its safety or recovery behavior, but replace the architecture-specific kernel
around it with the common TPU implementation.

\begin{table*}[t!]
\centering
\caption{\textbf{Absolute VBench scores} for Wan~2.2 T2V at 720p$\times$81f on
110 prompts, using VBench's per-dimension aggregates (higher is better).
The first five dimensions appear in Table~\ref{tab:safety_quality}.
VFA-style, FD++-style, and FA4-style (row) use TPU~v6e. Other methods use TPU7x.
$^\dagger$All 110 VFA-style videos have all-zero frames and are unscored.}
\label{tab:vbench_absolute}
\small
\resizebox{\textwidth}{!}{%
\begin{tabular}{lcccccc}
\toprule
Dimension & Online softmax & VFA-style & FD++-style & FA4-style (tile) & FA4-style (row) & \method{} \\ \midrule
Imaging quality & 0.6658 & --$^\dagger$ & 0.6668 & 0.6690 & 0.6672 & 0.6768 \\
Subject consistency & 0.9237 & --$^\dagger$ & 0.9231 & 0.9250 & 0.9224 & 0.9252 \\
Temporal flickering & 0.9693 & --$^\dagger$ & 0.9705 & 0.9692 & 0.9710 & 0.9697 \\
Motion smoothness & 0.9454 & --$^\dagger$ & 0.9453 & 0.9465 & 0.9452 & 0.9456 \\
Dynamic degree & 1.0000 & --$^\dagger$ & 0.9000 & 1.0000 & 0.9000 & 1.0000 \\ \midrule
Aesthetic quality & 0.6226 & --$^\dagger$ & 0.6240 & 0.6223 & 0.6216 & 0.6287 \\
Background consistency & 0.9178 & --$^\dagger$ & 0.9275 & 0.9264 & 0.9273 & 0.9206 \\
Overall consistency & 0.2644 & --$^\dagger$ & 0.2656 & 0.2629 & 0.2662 & 0.2651 \\
Appearance style & 0.2197 & --$^\dagger$ & 0.2201 & 0.2193 & 0.2192 & 0.2180 \\
Temporal style & 0.2397 & --$^\dagger$ & 0.2395 & 0.2376 & 0.2406 & 0.2400 \\
Object class & 0.7375 & --$^\dagger$ & 0.7188 & 0.7312 & 0.7875 & 0.7688 \\
Multiple objects & 0.6500 & --$^\dagger$ & 0.6312 & 0.6687 & 0.6375 & 0.7063 \\
Spatial relationship & 0.8370 & --$^\dagger$ & 0.8442 & 0.8568 & 0.8310 & 0.8356 \\
Human action & 0.6000 & --$^\dagger$ & 0.6000 & 0.7000 & 0.8000 & 0.6000 \\
Scene & 0.2875 & --$^\dagger$ & 0.2875 & 0.2188 & 0.1938 & 0.3000 \\
Color & 0.9609 & --$^\dagger$ & 0.8542 & 0.9554 & 0.8264 & 0.9844 \\
\bottomrule
\end{tabular}}
\end{table*}

\subsection{Implementation Details}

\paragraph{Tuned online softmax.}
For every KV block, the control computes the block row maximum, updates the
running maximum, and rescales the normalizer and output accumulator.  It uses
the same tiles, layouts, ring communication, BF16 inputs, and FP32 accumulators
as \method{}.  It is therefore the appropriate denominator for both latency
and numerical comparisons.

\paragraph{FlashDecoding++-style unified shift.}
FlashDecoding++ assigns all independently processed score partitions a unified
shift $\phi$, chosen from empirical score ranges, and falls back to the
synchronized online-softmax computation when the supported exponent interval
is violated.  Our port retains this shared-shift fast path, the runtime range
check, and exact recomputation, while excluding the paper's asynchronous CUDA
schedule and flat-GEMM optimization.  We calibrate $\phi$ on prompts and seeds
disjoint from evaluation.  The fast path accumulates $\exp(S_j-\phi)$ without
a running maximum or per-block rescaling. On a violation, the complete
affected row is recomputed with online softmax, and that cost is included in
latency.  This baseline therefore tests whether empirical calibration plus
recovery can replace a proof of safety.

\paragraph{VFA-style predicted maximum.}
VFA initializes each row's running maximum from signed absolute-maximum
representatives of the key blocks, processes the sink and local blocks with
ordinary online updates, and freezes the maximum for the remaining blocks.
Our port retains all three steps.  We map ``local'' to the key block covering
the same flattened spatiotemporal range as the video query block, and include
representative construction in latency.  We record any amount by which the
frozen maximum is exceeded but do not repair the primary result: VFA does not
specify a recovery path, and adding one would define a different method.  The
baseline therefore measures both the performance and the safety consequences
of transferring its approximate initialization and freezing rule to long
video sequences.

\paragraph{FA4-style conditional rescaling.}
FlashAttention-4 retains the exact row-maximum reduction for every block.  For
each query row, let $\Delta_j$ be the increase from the retained reference to
the candidate maximum for block $j$.  The reference and accumulators are
rescaled only when $\Delta_j>\tau$, with $\tau=8$ in base-2 exponent units.
Otherwise, the row continues to accumulate under the previous reference.  The
published rule is therefore row-wise.  Its Blackwell kernel coalesces only the
\emph{execution} of correction: one thread owns each row, and a 32-thread warp
runs the correction path when any of its rows needs it, using a unit scale for
the other rows.  This is neither an independently branched implementation nor
a single decision for the full 128-row query tile.

TPUs have no direct analogue of this CUDA warp-level control.  We therefore
evaluate two Pallas mappings: a row-wise variant that follows the published
threshold semantics, and a tile-coalesced variant that promotes an any-row
trigger to the full query tile.  We report both variants separately in the
performance and VBench comparisons, labeling their control granularity as
row or tile.  Both mappings retain the
per-block row-maximum reduction and isolate conditional rescaling from
FlashAttention-4's tensor-memory pipeline, warp specialization, and
polynomial exponential approximation.

\paragraph{Uncertified control.}
Replaying the rejected heads with an uncertified fixed shift shows the failures
the certificate prevents (Table~\ref{tab:fallback_accounting}b).

\subsection{Safety and Numerical Measurements}
\label{app:safety_measurements}

\begin{table*}[t!]
\centering
\caption{\textbf{Paired VBench differences from online softmax} with 95\%
percentile-bootstrap confidence intervals over prompts.
$n$ counts prompts scored for both methods.}
\label{tab:vbench_delta_ci}
\small
\resizebox{\textwidth}{!}{%
\begin{tabular}{lccccc}
\toprule
Dimension & FD++-style & FA4-style (tile) & FA4-style (row) & \method{} & $n$ \\ \midrule
Imaging quality & $+0.0011$ [$-0.0158$, $+0.0198$] & $+0.0032$ [$-0.0061$, $+0.0149$] & $+0.0014$ [$-0.0124$, $+0.0181$] & $+0.0110$ [$+0.0015$, $+0.0239$] & 10 \\
Subject consistency & $-0.0005$ [$-0.0106$, $+0.0096$] & $+0.0013$ [$-0.0029$, $+0.0061$] & $-0.0013$ [$-0.0082$, $+0.0068$] & $+0.0015$ [$-0.0043$, $+0.0076$] & 10 \\
Temporal flickering & $+0.0012$ [$-0.0005$, $+0.0035$] & $-0.0001$ [$-0.0004$, $+0.0002$] & $+0.0017$ [$-0.0001$, $+0.0041$] & $+0.0004$ [$+0.0000$, $+0.0010$] & 10 \\
Motion smoothness & $-0.0001$ [$-0.0028$, $+0.0026$] & $+0.0010$ [$-0.0006$, $+0.0032$] & $-0.0003$ [$-0.0031$, $+0.0032$] & $+0.0002$ [$-0.0010$, $+0.0016$] & 10 \\
Dynamic degree & $-0.1000$ [$-0.3000$, $+0.0000$] & $+0.0000$ [$+0.0000$, $+0.0000$] & $-0.1000$ [$-0.3000$, $+0.0000$] & $+0.0000$ [$+0.0000$, $+0.0000$] & 10 \\
\bottomrule
\end{tabular}}
\end{table*}

\begin{table*}[t]
\centering
\caption{\textbf{Parallelism and attention-kernel ablation on TPU7x-8 (Wan~2.2, 720p$\times$81f, $S=75{,}600, H=40, D=128$).}
Online softmax and \method{} use independently tuned tiles for each setting.
All settings use the same eight TensorCores (four chips). The four
context-parallel columns disable CFG and use eight-way context parallelism
(CFG$=$1, Ulysses$\times$Ring$=8$), whereas Figure~\ref{fig:main_eval_all} and
Tables~\ref{tab:kernel_latency_absolute}--\ref{tab:dit_latency_absolute} use
CFG$=$2$\times$Ring$=$4. Absolute latencies are therefore not comparable across them.}
\label{tab:parallelism_ablation}
\footnotesize
\setlength{\tabcolsep}{4pt}
\renewcommand{\arraystretch}{1.12}
\begin{tabular}{l c c c c c}
\toprule
\textbf{Parallelism} & \textbf{CP=1, DP=8} & \textbf{Ulysses=8, Ring=1} & \textbf{Ulysses=4, Ring=2} & \textbf{Ulysses=2, Ring=4} & \textbf{Ulysses=1, Ring=8} \\
\textit{Kernel family} & \textit{Splash} & \textit{Splash + A2A} & \textit{2D Ring} & \textit{2D Ring} & \textit{Pure Ring} \\
\textit{Local shard $(H_{\mathrm{loc}}, S_{\mathrm{shard}})$} & $(40,\,75{,}600)$ & $(5,\,75{,}600)$ & $(10,\,37{,}800)$ & $(20,\,18{,}900)$ & $(40,\,9{,}450)$ \\
\midrule
\multicolumn{6}{l}{\textbf{Attention Kernel Microbenchmark (ms)}} \\
Online softmax & 185.85 & 31.62 & 32.95 & 28.26 & 27.67 \\
\method{} & \textbf{173.32} & \textbf{29.54} & \textbf{29.74} & \textbf{25.06} & \textbf{24.66} \\
Latency change & $\mathbf{-6.7\%}$ & $\mathbf{-6.6\%}$ & $\mathbf{-9.8\%}$ & $\mathbf{-11.3\%}$ & $\mathbf{-10.9\%}$ \\
\midrule
\multicolumn{6}{l}{\textbf{DiT Denoising Step Latency (s/step)}} \\
Online softmax & 10.041 & 1.519 & 1.595 & 1.430 & 1.415 \\
\method{} & \textbf{9.544} & \textbf{1.474} & \textbf{1.504} & \textbf{1.351} & \textbf{1.308} \\
Latency change & $\mathbf{-5.0\%}$ & $\mathbf{-2.9\%}$ & $\mathbf{-5.7\%}$ & $\mathbf{-5.6\%}$ & $\mathbf{-7.5\%}$ \\
\bottomrule
\end{tabular}
\end{table*}

\paragraph{Safe-condition rates.}
The rates in Table~\ref{tab:safety_quality} use each method's native decision unit and are therefore not directly comparable.
For \method{}, the rate is the fraction of (call, head) pairs that pass certification in the 40-step 720p run behind Table~\ref{tab:fallback_accounting} and Figure~\ref{fig:fallback_scaling}, over 25{,}600 calls at each of the five 720p sequence lengths (39.6K--184K tokens).
For FD++, the rate is the fraction of ring steps that avoid recomputation. At 720p$\times$81f, 691 of 7{,}680 steps trigger recomputation.
For VFA, we sample one attention call per denoising step and report the fraction of (row, head) pairs whose frozen $m_{\mathrm{init}}$ covers the true row maximum.
This coverage ranges from 23.8--54.6\% over 40 steps ($3.2\!\times\!10^7$ sampled pairs) as the maximum logit grows from 7.1 to 70.7 in base-2 units.
FA4 is unconditionally correct and requires no certificate.

\paragraph{Attention-output error.}
We compare each method with online softmax on identical Q/K/V captured from a Wan~2.2 720p$\times$81f denoising step (one ring shard, $(1,20,37800,128)$ BF16), using each method's end-to-end tile.
The online kernel itself differs from an exact FP32 reference by $6.09\!\times\!10^{-3}$ in relative $\ell_2$ at this shape.
The common maximum error of $3.1\!\times\!10^{-2}$ is one BF16 ULP at the output magnitude and therefore does not discriminate among the non-online variants.
Table~\ref{tab:safety_quality} summarizes the full-tile-control FA4-style
TPU adaptation, labeled FA4-style (tile). Its row-wise counterpart has relative $\ell_2$ error
$2.30\!\times\!10^{-3}$ and maximum absolute error $6.3\!\times\!10^{-2}$.
Both variants are reported separately in the performance comparisons
(Figure~\ref{fig:main_eval_all} and Tables~\ref{tab:kernel_latency_absolute}
and~\ref{tab:dit_latency_absolute}) and VBench results
(Tables~\ref{tab:vbench_absolute} and~\ref{tab:vbench_delta_ci}).

\begin{figure*}[t!]
  \centering
  \includegraphics[width=\textwidth]{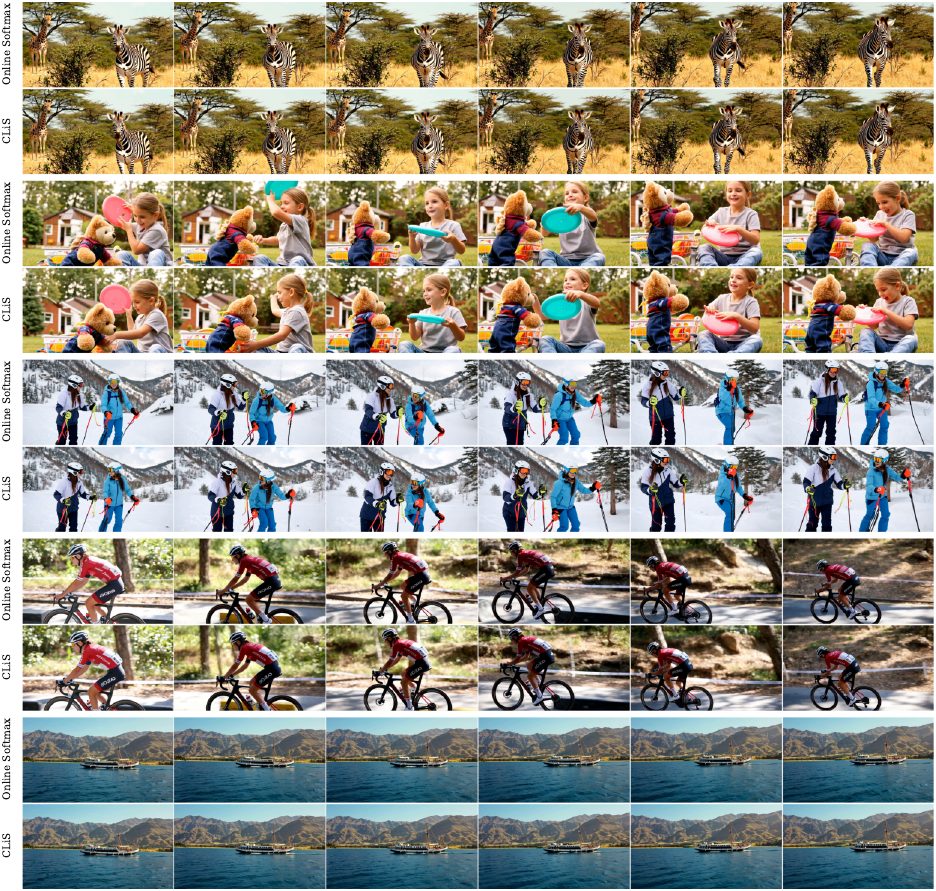}
  \caption{Visual examples of video generation on Wan~2.2 (720p$\times$81f, 40 denoising steps, with frames 0, 16, 32, 48, 64, 80 shown) across VBench prompts comparing tuned online softmax and \method{}.}
  \label{fig:vbench_visual_comparison}
\end{figure*}

\paragraph{VBench.}
We generate one Wan~2.2 T2V video per prompt of a 110-prompt stratified VBench subset (720p$\times$81f, 40 UniPC~\citep{zhao2023unipc} steps, the same random seed, flow shift 3.0, BF16 with FP8 attention and linears disabled) for each method, and score all 16 VBench dimensions with the upstream VBench evaluator.
Each dimension is scored on the prompts tagged with it, 10 per dimension. Color additionally drops videos in which the detector finds no target object.
Table~\ref{tab:vbench_absolute} reports VBench's aggregate scores, Table~\ref{tab:vbench_delta_ci} the paired deltas behind Table~\ref{tab:safety_quality}, and Figure~\ref{fig:vbench_visual_comparison} provides side-by-side visual comparisons across representative VBench prompts.
Online softmax, \method{}, and FA4-style (tile) videos were generated on TPU7x, and FD++-style, FA4-style (row), and VFA-style videos on TPU~v6e. VBench scores each video independently, and generating the same method on the other TPU changes rounding, not the method.
Because online softmax, \method{}, FD++-style, and FA4-style all compute the same attention in exact arithmetic, their deltas measure how far BF16 rounding, amplified over the 40-step trajectory, moves the sampled video. Differences of this size between two exact methods set the noise floor against which Table~\ref{tab:safety_quality} should be read.
Every VFA-style video decodes to all-zero frames and is therefore not scored.

\begin{table*}[p]
\centering
\caption{\textbf{Absolute attention-kernel latency (ms)} for the 40 workloads
in Figure~\ref{fig:main_eval_all}. Values are means.
All runs use eight TensorCores with CFG$=$2$\times$Ring$=$4 (Ulysses$=$1).
FD++ and FA4 denote the style baselines.
$^{\dagger}$Corrupted output.}
\label{tab:kernel_latency_absolute}
\footnotesize
\setlength{\tabcolsep}{4.5pt}
\renewcommand{\arraystretch}{1.04}
\begin{tabular}{r r r r r r r r}
\toprule
Frames & Tokens & Online & FD++ & VFA & FA4 tile & FA4 row & \method{} \\
\midrule
\multicolumn{8}{l}{\textbf{Wan~2.2, TPU~v6e, 480p}} \\
41 & 17,160 & 4.089 & 4.809 & 6.931$^{\dagger}$ & 7.624 & 4.086 & 3.344 \\
81 & 32,760 & 14.156 & 18.320 & 12.926$^{\dagger}$ & 21.098 & 14.152 & 11.012 \\
121 & 48,360 & 30.666 & 37.374 & 26.123$^{\dagger}$ & 47.123 & 30.645 & 24.169 \\
161 & 63,960 & 52.009 & 61.793 & 43.164$^{\dagger}$ & 64.258 & 52.045 & 40.841 \\
201 & 79,560 & 80.151 & 97.233 & 64.993 & 97.864 & 80.280 & 61.110 \\
\midrule
\multicolumn{8}{l}{\textbf{Wan~2.2, TPU~v6e, 720p}} \\
41 & 39,600 & 20.986 & 24.855 & 17.963$^{\dagger}$ & 42.107 & 20.783 & 16.819 \\
81 & 75,600 & 73.002 & 89.220 & 60.533$^{\dagger}$ & 90.772 & 73.034 & 57.629 \\
121 & 111,600 & 159.764 & 186.821 & 126.678$^{\dagger}$ & 188.396 & 157.149 & 122.610 \\
161 & 147,600 & 274.548 & 325.357 & 216.361$^{\dagger}$ & 323.929 & 274.832 & 211.768 \\
201 & 183,600 & 419.815 & 496.323 & 327.618$^{\dagger}$ & 495.442 & 414.598 & 320.727 \\
\midrule
\multicolumn{8}{l}{\textbf{Wan~2.2, TPU7x, 480p}} \\
41 & 17,160 & 3.231 & 3.552 & 3.004$^{\dagger}$ & 3.446 & 3.214 & 2.754 \\
81 & 32,760 & 10.834 & 12.145 & 9.172$^{\dagger}$ & 11.207 & 10.827 & 8.472 \\
121 & 48,360 & 23.413 & 25.905 & 19.695$^{\dagger}$ & 24.245 & 23.356 & 18.336 \\
161 & 63,960 & 40.277 & 39.525 & 32.358$^{\dagger}$ & 85.607 & 40.324 & 31.262 \\
201 & 79,560 & 54.483 & 60.511 & 49.217 & 150.986 & 54.604 & 47.605 \\
\midrule
\multicolumn{8}{l}{\textbf{Wan~2.2, TPU7x, 720p}} \\
41 & 39,600 & 13.863 & 15.399 & 12.782$^{\dagger}$ & 36.532 & 13.864 & 12.302 \\
81 & 75,600 & 48.894 & 55.156 & 44.813$^{\dagger}$ & 148.243 & 49.270 & 43.712 \\
121 & 111,600 & 105.769 & 117.437 & 95.412$^{\dagger}$ & 261.834 & 106.638 & 92.399 \\
161 & 147,600 & 210.893 & 205.882 & 166.704$^{\dagger}$ & 429.516 & 210.815 & 162.913 \\
201 & 183,600 & 286.316 & 315.458 & 256.481$^{\dagger}$ & 683.688 & 288.964 & 245.522 \\
\midrule
\multicolumn{8}{l}{\textbf{LTX-2, TPU~v6e, 960p}} \\
81 & 17,160 & 3.331 & 4.062 & 5.745 & 6.489 & 3.478 & 2.715 \\
161 & 32,760 & 11.607 & 14.892 & 10.609$^{\dagger}$ & 17.291 & 11.601 & 8.822 \\
241 & 48,360 & 24.538 & 29.641 & 20.910$^{\dagger}$ & 37.707 & 24.522 & 19.083 \\
321 & 63,960 & 41.601 & 48.902 & 34.534$^{\dagger}$ & 51.396 & 41.632 & 32.696 \\
401 & 79,560 & 64.113 & 76.991 & 51.996$^{\dagger}$ & 78.271 & 64.207 & 48.900 \\
\midrule
\multicolumn{8}{l}{\textbf{LTX-2, TPU~v6e, 1440p}} \\
81 & 39,600 & 16.793 & 19.733 & 14.388$^{\dagger}$ & 33.689 & 16.632 & 13.055 \\
161 & 75,600 & 58.398 & 70.617 & 48.425$^{\dagger}$ & 72.615 & 58.419 & 46.125 \\
241 & 111,600 & 127.808 & 148.051 & 101.338$^{\dagger}$ & 150.723 & 125.716 & 98.126 \\
321 & 147,600 & 219.500 & 257.058 & 173.001$^{\dagger}$ & 258.949 & 219.732 & 169.437 \\
401 & 183,600 & 335.787 & 393.414 & 262.044$^{\dagger}$ & 396.267 & 331.613 & 256.590 \\
\midrule
\multicolumn{8}{l}{\textbf{LTX-2, TPU7x, 960p}} \\
81 & 17,160 & 2.609 & 2.878 & 2.446 & 2.797 & 2.615 & 2.260 \\
161 & 32,760 & 8.625 & 9.648 & 7.337$^{\dagger}$ & 8.918 & 8.622 & 6.801 \\
241 & 48,360 & 18.721 & 20.827 & 15.823$^{\dagger}$ & 19.484 & 18.730 & 14.654 \\
321 & 63,960 & 28.120 & 31.074 & 25.712$^{\dagger}$ & 65.615 & 28.179 & 25.024 \\
401 & 79,560 & 43.124 & 48.246 & 39.284$^{\dagger}$ & 119.369 & 43.598 & 38.160 \\
\midrule
\multicolumn{8}{l}{\textbf{LTX-2, TPU7x, 1440p}} \\
81 & 39,600 & 11.141 & 13.915 & 10.251$^{\dagger}$ & 13.197 & 11.139 & 10.091 \\
161 & 75,600 & 39.171 & 44.231 & 35.899$^{\dagger}$ & 118.927 & 39.478 & 35.073 \\
241 & 111,600 & 84.743 & 94.343 & 76.590$^{\dagger}$ & 210.342 & 85.611 & 73.919 \\
321 & 147,600 & 168.894 & 165.060 & 134.189$^{\dagger}$ & 345.569 & 168.920 & 130.405 \\
401 & 183,600 & 229.212 & 252.108 & 205.362$^{\dagger}$ & 546.896 & 231.421 & 196.345 \\
\bottomrule
\end{tabular}
\end{table*}

\begin{table*}[p]
\centering
\caption{\textbf{Absolute DiT-step latency (s)} for the 40 workloads
in Figure~\ref{fig:main_eval_all}. Values are means.
All runs use eight TensorCores with CFG$=$2$\times$Ring$=$4 (Ulysses$=$1).
FD++ and FA4 denote the style baselines.
$^{\dagger}$Corrupted output.}
\label{tab:dit_latency_absolute}
\footnotesize
\setlength{\tabcolsep}{4.5pt}
\renewcommand{\arraystretch}{1.04}
\begin{tabular}{r r r r r r r r}
\toprule
Frames & Tokens & Online & FD++ & VFA & FA4 tile & FA4 row & \method{} \\
\midrule
\multicolumn{8}{l}{\textbf{Wan~2.2, TPU~v6e, 480p}} \\
41 & 17,160 & 0.606 & 0.635 & 0.768$^{\dagger}$ & 0.754 & 0.606 & 0.574 \\
81 & 32,760 & 1.159 & 1.325 & 1.233$^{\dagger}$ & 1.445 & 1.159 & 1.039 \\
121 & 48,360 & 2.055 & 2.306 & 1.873$^{\dagger}$ & 2.696 & 2.053 & 1.797 \\
161 & 63,960 & 3.159 & 3.527 & 2.807$^{\dagger}$ & 3.643 & 3.160 & 2.719 \\
201 & 79,560 & 4.550 & 5.203 & 3.935 & 5.254 & 4.555 & 3.799 \\
\midrule
\multicolumn{8}{l}{\textbf{Wan~2.2, TPU~v6e, 720p}} \\
41 & 39,600 & 1.541 & 1.685 & 1.432$^{\dagger}$ & 2.359 & 1.533 & 1.376 \\
81 & 75,600 & 4.211 & 4.834 & 3.721$^{\dagger}$ & 4.921 & 4.213 & 3.607 \\
121 & 111,600 & 8.340 & 9.369 & 6.985$^{\dagger}$ & 9.494 & 8.235 & 6.890 \\
161 & 147,600 & 13.506 & 15.480 & 11.216$^{\dagger}$ & 15.506 & 13.516 & 11.042 \\
201 & 183,600 & 19.855 & 22.811 & 16.154$^{\dagger}$ & 22.900 & 19.648 & 15.986 \\
\midrule
\multicolumn{8}{l}{\textbf{Wan~2.2, TPU7x, 480p}} \\
41 & 17,160 & 0.305 & 0.312 & 0.296$^{\dagger}$ & 0.418 & 0.304 & 0.284 \\
81 & 32,760 & 0.735 & 0.746 & 0.671$^{\dagger}$ & 1.052 & 0.735 & 0.642 \\
121 & 48,360 & 1.376 & 1.406 & 1.231$^{\dagger}$ & 2.791 & 1.378 & 1.175 \\
161 & 63,960 & 2.199 & 2.221 & 1.872$^{\dagger}$ & 4.024 & 2.201 & 1.839 \\
201 & 79,560 & 2.898 & 3.203 & 2.691 & 6.745 & 2.903 & 2.629 \\
\midrule
\multicolumn{8}{l}{\textbf{Wan~2.2, TPU7x, 720p}} \\
41 & 39,600 & 0.936 & 1.014 & 0.893$^{\dagger}$ & 1.838 & 0.936 & 0.872 \\
81 & 75,600 & 2.633 & 2.943 & 2.472$^{\dagger}$ & 6.597 & 2.646 & 2.430 \\
121 & 111,600 & 5.203 & 5.806 & 4.797$^{\dagger}$ & 11.448 & 5.237 & 4.686 \\
161 & 147,600 & 9.710 & 9.790 & 7.999$^{\dagger}$ & 18.488 & 9.715 & 7.809 \\
201 & 183,600 & 13.048 & 14.531 & 11.867$^{\dagger}$ & 28.796 & 13.160 & 11.474 \\
\midrule
\multicolumn{8}{l}{\textbf{LTX-2, TPU~v6e, 960p}} \\
81 & 17,160 & 0.715 & 0.745 & 0.767 & 0.806 & 0.715 & 0.693 \\
161 & 32,760 & 1.378 & 1.542 & 1.521$^{\dagger}$ & 1.647 & 1.379 & 1.274 \\
241 & 48,360 & 2.309 & 2.568 & 2.158$^{\dagger}$ & 2.643 & 2.309 & 2.063 \\
321 & 63,960 & 3.506 & 3.881 & 3.159$^{\dagger}$ & 4.117 & 3.508 & 3.099 \\
401 & 79,560 & 5.026 & 5.678 & 4.441$^{\dagger}$ & 5.887 & 5.028 & 4.310 \\
\midrule
\multicolumn{8}{l}{\textbf{LTX-2, TPU~v6e, 1440p}} \\
81 & 39,600 & 1.736 & 1.920 & 1.707$^{\dagger}$ & 1.968 & 1.729 & 1.599 \\
161 & 75,600 & 4.517 & 5.139 & 4.026$^{\dagger}$ & 5.380 & 4.517 & 3.952 \\
241 & 111,600 & 8.709 & 9.759 & 7.448$^{\dagger}$ & 10.144 & 8.612 & 7.367 \\
321 & 147,600 & 13.809 & 15.751 & 11.562$^{\dagger}$ & 15.711 & 13.817 & 11.399 \\
401 & 183,600 & 20.240 & 23.190 & 16.702$^{\dagger}$ & 23.805 & 20.047 & 17.203 \\
\midrule
\multicolumn{8}{l}{\textbf{LTX-2, TPU7x, 960p}} \\
81 & 17,160 & 0.330 & 0.330 & 0.327 & 0.393 & 0.331 & 0.311 \\
161 & 32,760 & 0.761 & 0.760 & 0.711$^{\dagger}$ & 1.023 & 0.762 & 0.687 \\
241 & 48,360 & 1.433 & 1.427 & 1.351$^{\dagger}$ & 2.282 & 1.434 & 1.250 \\
321 & 63,960 & 2.042 & 2.171 & 1.937$^{\dagger}$ & 3.863 & 2.048 & 1.925 \\
401 & 79,560 & 2.986 & 3.213 & 2.813$^{\dagger}$ & 6.292 & 3.002 & 2.758 \\
\midrule
\multicolumn{8}{l}{\textbf{LTX-2, TPU7x, 1440p}} \\
81 & 39,600 & 0.955 & 1.003 & 0.917$^{\dagger}$ & 1.787 & 0.955 & 0.922 \\
161 & 75,600 & 2.665 & 2.885 & 2.520$^{\dagger}$ & 6.425 & 2.676 & 2.498 \\
241 & 111,600 & 5.196 & 5.624 & 4.836$^{\dagger}$ & 10.599 & 5.234 & 4.689 \\
321 & 147,600 & 9.656 & 9.536 & 8.075$^{\dagger}$ & 17.720 & 9.659 & 7.890 \\
401 & 183,600 & 12.916 & 14.041 & 11.812$^{\dagger}$ & 27.445 & 13.031 & 11.495 \\
\bottomrule
\end{tabular}
\end{table*}

\subsection{Why GPU Kernels Do Not Transfer Directly to TPUs}

The limitation is not that the GPU algorithms are invalid.  Their complete
implementations are co-designed with a different execution model, so neither
their source code nor their performance assumptions carry over unchanged.

\paragraph{Different execution substrate.}
GPU kernels are organized around CUDA thread blocks, warps, shared memory, and
architecture-specific asynchronous tensor-core pipelines
\citep{nvidia2022h100,nvidia2024blackwell}.  TPU kernels are written in Pallas,
lowered through Mosaic/XLA, and compiled into VLIW LLO bundles that coordinate
the MXU, XLU, and VPU through a shared vector-register file
\citep{jouppi2021ten,jax2026pallasdesign}.  GPU instructions, synchronization,
and memory primitives therefore have no direct TPU implementation.

\paragraph{Different dataflow and control costs.}
FlashDecoding++ couples its shift heuristic with GPU-specific flat-GEMM and
asynchronous execution.  FlashAttention-4 relies on a Blackwell-oriented
pipeline and memory hierarchy.  VFA's sink/local block reordering can conflict
with the arrival order of ring-sharded KV blocks on TPU, requiring prefetching
or buffering.  Moreover, a row-wise branch that is cheap on a GPU may compile
to masked work or execute both paths on a TPU, so whether it removes work must
be verified in LLO.

\paragraph{Different remaining critical paths.}
After adaptation, FA4-style execution still performs a row-maximum reduction
on every block, preserving the XLU-to-vector dependency that dominates the TPU
stabilization chain.  FD++-style execution adds range checks and potentially
expensive recomputation.  VFA removes more work, but its approximate shift can
under-estimate long-video logits and leave the safe exponent range.  As a
result, the adaptations may be correct yet slower, or fast yet unsafe.  Our controlled
ports retain each paper's softmax-side idea on a common TPU kernel, allowing us
to measure this transfer gap without claiming to reproduce the original GPU
systems.

\section{Parallelism and Attention-Kernel Ablation}
\label{app:parallelism_ablation}

To verify that \method{} provides consistent speedups across distinct attention
kernel implementations and context-parallelism decompositions rather than
benefiting only a single sharding point, Table~\ref{tab:parallelism_ablation}
evaluates five 8-TensorCore configurations on TPU7x-8 (four chips) for Wan~2.2 at 720p$\times$81f
($S=75{,}600, H=40, D=128$): data-parallel execution without context
parallelism (\texttt{CP=1, DP=8}) using standalone Splash attention,
pure Ulysses head-sharded context parallelism (\texttt{Ulysses=8, Ring=1})
combining Splash attention with all-to-all collectives, two hybrid 2D
context-parallel layouts (\texttt{Ulysses=4, Ring=2} and \texttt{Ulysses=2, Ring=4})
combining ring attention with all-to-all collectives, and
pure sequence-sharded ring attention (\texttt{Ulysses=1, Ring=8}).
The main evaluation (Figure~\ref{fig:main_eval_all} and
Tables~\ref{tab:kernel_latency_absolute}--\ref{tab:dit_latency_absolute})
uses the same eight TensorCores as two classifier-free-guidance groups of
four-way ring attention (\texttt{CFG=2, Ring=4, Ulysses=1}). To expose more
Ulysses--Ring factorizations, the context-parallel settings here instead
disable CFG and use eight-way context parallelism
(\texttt{CFG=1}, \texttt{Ulysses}$\times$\texttt{Ring}$=8$). In this layout,
each TensorCore holds half the attention work, and each DiT step runs one guidance pass
instead of two, so absolute latencies are lower than the 720p$\times$81f
entries in Tables~\ref{tab:kernel_latency_absolute}--\ref{tab:dit_latency_absolute}
and are not directly comparable.
For every setting, we sweep candidate tile shapes independently for both
online softmax and \method{} and report each method at its fastest tile.
Across all five configurations, \method{} consistently reduces attention-kernel
latency by $6.6\text{--}11.3\%$ ($1.07\text{--}1.13\times$) and full 40-block
DiT denoising-step latency by $2.9\text{--}7.5\%$ ($1.03\text{--}1.08\times$),
with pure ring attention (\texttt{Ulysses=1, Ring=8}) achieving the
lowest overall kernel ($24.66$~ms) and DiT-step ($1.308$~s/step) latency.

\section{Absolute Latency Measurements}
\label{app:absolute_latency}

As shown in Tables~\ref{tab:kernel_latency_absolute} and~\ref{tab:dit_latency_absolute},
we report absolute attention-kernel and DiT-step latencies for all workloads
in Figure~\ref{fig:main_eval_all}. The tables use the same recorded means
and online-softmax normalization controls as the figure.

\end{document}